\documentclass[letterpaper,journal]{IEEEtran}
\usepackage{amsmath,amsfonts}
\usepackage{algorithmic}
\usepackage{algorithm}
\usepackage{array}
\usepackage[caption=false,font=normalsize,labelfont=sf,textfont=sf]{subfig}
\usepackage{textcomp}
\usepackage{stfloats}
\usepackage{url}
\usepackage{verbatim}
\usepackage{graphicx}
\usepackage{cite}
\usepackage{epsfig}
\usepackage{booktabs}
\usepackage{multirow}
\usepackage{makecell}
\usepackage{tabularx}
\usepackage[hidelinks]{hyperref}
\def\BibTeX{{\rm B\kern-.05em{\sc i\kern-.025em b}\kern-.08em
T\kern-.1667em\lower.7ex\hbox{E}\kern-.125emX}}
\usepackage{balance}

\newcommand{\IEEEacceptednotice}{%
  \begingroup
  \setlength{\fboxsep}{6pt}%
  \noindent\fbox{%
    \begin{minipage}{\dimexpr\textwidth-2\fboxsep-2\fboxrule\relax}
      \footnotesize
      \textbf{IEEE Copyright Notice}\\
      This is the accepted manuscript version of an article accepted for publication in \textit{IEEE Journal of Oceanic Engineering}. \copyright~2026 IEEE. Personal use of this material is permitted. Permission from IEEE must be obtained for all other uses, in any current or future media, including reprinting/republishing this material for advertising or promotional purposes, creating new collective works, for resale or redistribution to servers or lists, or reuse of any copyrighted component of this work in other works.
    \end{minipage}%
  }\par\vspace{1.2em}
  \endgroup
}

\begin{document}

\title{Raspi$^2$USBL: An open-source Raspberry Pi-Based Passive Inverted Ultra-Short Baseline Positioning System for Underwater Robotics}

\author{Jin Huang,
  \IEEEmembership{Graduate Student Member, IEEE},
  Yingqiang Wang,
  \IEEEmembership{Member, IEEE},
  Ying Chen
  \thanks{Corresponding authors: Y. Wang and Y. Chen.}
  \thanks{J. Huang and Y. Chen are with the State Key Laboratory of Ocean Sensing, Ocean College of Zhejiang University, Zhoushan, 316021, China (e-mail: jin.huang@zju.edu.cn; ychen@zju.edu.cn).}
  \thanks{Y. Wang is with the School of Oceanography, Shanghai Jiao Tong University, Shanghai, 200240, China (e-mail: wyingqiang@sjtu.edu.cn).}
  \thanks{This work was supported in part by the National Natural Science Foundation of China under Grant 52501422, in part by the Natural Science Foundation of Shanghai under Grant 25ZR1402253, and in part by the Postdoctoral Fellowship Program of CPSF under Grant GZC20250936.}
}
\markboth{}
{Jin Huang \MakeLowercase{\textit{et al.}}: Raspi$^2$USBL: An open-source Raspberry Pi-Based Passive Inverted Ultra-Short Baseline Positioning System for Underwater Robotics}


\IEEEaftertitletext{\IEEEacceptednotice}

\maketitle

\begin{abstract}
  Precise underwater positioning remains a fundamental challenge for underwater robotics since global navigation satellite system (GNSS) signals cannot penetrate the sea surface.
  This paper presents Raspi$^2$USBL, a Raspberry Pi-based passive inverted ultra-short baseline (piUSBL) positioning system designed to provide a low-cost, accessible, and reproducible solution for underwater robotic research.
  The system comprises a passive acoustic receiver and an active beacon.
  The receiver adopts a modular hardware architecture that integrates a hydrophone array, a multichannel preamplifier, an oven-controlled crystal oscillator (OCXO), a Raspberry Pi 5, and an MCC-series data acquisition (DAQ) board.
  Apart from the Raspberry Pi, OCXO, and MCC board, the beacon comprises an impedance-matching network, a power amplifier, and a transmitting transducer.
  An open-source C++ software framework provides high-precision clock synchronization and triggering for one-way travel-time (OWTT) messaging, while performing real-time signal processing, including matched filtering, array beamforming, and adaptive gain control, to estimate the time of flight (TOF) and direction of arrival (DOA) of received signals.
  The Raspi$^2$USBL system was experimentally validated in an anechoic tank, freshwater lake, and open-sea trials.
  Results demonstrate a slant-range accuracy better than 0.1\%, a bearing accuracy within 0.1$^\circ$, and stable performance over operational distances up to 1.3 km.
  These findings confirm that low-cost, system-level reproducible hardware can deliver research-grade underwater positioning accuracy.
  By releasing the software framework as open-source and providing a reproducible hardware architecture, Raspi$^2$USBL offers a unified reference platform that lowers the entry barrier for underwater robotics laboratories, fosters reproducibility, and promotes collaborative innovation in underwater acoustic navigation and swarm robotics.
\end{abstract}


\section{Introduction}
\IEEEPARstart{D}{ue} to the inability of electromagnetic waves to penetrate underwater, the Global Navigation Satellite System (GNSS) is unavailable in such environments \cite{zhangAutonomousUnderwaterVehicle2023}.
To overcome this limitation, various underwater positioning systems have been developed, broadly classified into inertial/dead reckoning, acoustic, and geophysical methods \cite{paullAUVNavigationLocalization2014a}.
However, inertial navigation systems (INS) suffer from cumulative drift over time, which leads to significant errors in long-term navigation \cite{millerAutonomousUnderwaterVehicle2010}.
Geophysical methods, such as magnetic, gravity, and terrain-aided navigation, require extensive prior knowledge of the environment and are often constrained by the availability of distinct features \cite{maUnderwaterMultizonotopeTerrainaided2025}.
Among these, acoustic methods are the most widely used due to their long-range capabilities and high accuracy \cite{zhaoReviewUnderwaterMultisource2023}.

Acoustic positioning systems are generally classified into three main types: long baseline (LBL), short baseline (SBL), and ultra-short baseline (USBL) systems \cite{sekimoriBearingElevationDepth2024}.
Among these, the USBL system is the most compact and portable, rendering it particularly suitable for diverse marine applications, especially for autonomous underwater vehicles (AUVs) \cite{guoRobustAttitudeEstimation2023}.
To meet the control and swarm research demands of AUVs, the USBL has evolved into several variants, including the inverted USBL (iUSBL) \cite{wangResearchPassivePositioning2025}, passive inverted USBL (piUSBL) \cite{rypkemaPassiveInvertedUltraShort2019}, and hybrid baseline (HBL) \cite{wangOneWayTravelTimeHybridBaseline2025} positioning systems.
In contrast to the traditional USBL system, the iUSBL places the transducer array on the AUV and the beacon on the surface platform.
This configuration enables the vehicle to measure its relative position to the beacon directly using the two-way travel time (TWTT) method for ranging \cite{sunInvertedUltrashortBaseline2019}.
The piUSBL system further simplifies ranging by synchronizing the clocks of the beacon and AUVs, thereby enabling one-way travel time (OWTT) measurements.
This approach markedly simplifies the system architecture and enhances applicability to AUV swarm positioning.
Compared with the USBL system, the piUSBL offers lower cost, higher efficiency, greater scalability, and reduced power consumption, making it more suitable for AUV swarm applications \cite{rypkemaSynchronousClockRangeAngleRelative2022}.

The evolution from conventional USBL to piUSBL systems has significantly improved scalability, efficiency, and suitability for AUV swarm positioning applications.
However, several critical limitations remain \cite{wangPassiveInvertedUltraShort2022}, posing substantial barriers to entry for new researchers.
First, the development of an acoustic positioning system is inherently complex, as it requires the integration of multidisciplinary expertise spanning underwater acoustics, electronics, and signal processing.
The high degree of technical coupling among these domains results in a steep learning curve and substantial engineering effort for system developers.
Second, although advanced signal processing algorithms and integrated navigation techniques have been increasingly adopted in recent years, most existing systems remain proprietary and closed-source.
This restricts access to raw acoustic and navigation data, thereby preventing researchers from validating, improving, or extending existing methods and ultimately slowing technological progress in the field.
Third, despite the architectural simplification introduced by piUSBL systems through the use of OWTT messaging, practical implementations still face major challenges in achieving stable signal acquisition and robust real-time processing during field deployments.
Finally, the overall development and experimental validation of acoustic systems are constrained by high costs, long development cycles, and limited access to specialized testing facilities such as anechoic water tanks.
These factors collectively restrict broader participation, hinder iterative experimentation, and delay the dissemination of innovative underwater localization in marine robotics.

To address these challenges, this paper introduces Raspi$^2$USBL, a Raspberry Pi-based passive inverted USBL positioning system with an open-source software framework and reproducible hardware architecture, designed to provide a low-cost, accessible, and extensible platform for underwater robotics research.
The significance of the system is threefold:
\begin{itemize}
  \item \textbf{Lowering the entry barrier for underwater robotics research.}
    By providing an open-source implementation that is both accessible and modifiable, Raspi$^2$USBL reduces the financial and technical burdens of developing underwater acoustic positioning systems, one of the key enabling technologies in underwater robotics.
    This democratization of access allows a wider range of researchers and institutions to engage in this field, thereby accelerating innovation and promoting the widespread adoption of underwater robotics in scientific applications such as ocean observation.
  \item \textbf{Enabling multi-vehicle swarm applications.}
    By employing a broadcast-based OWTT messaging architecture, the proposed system allows a single beacon to simultaneously localize and transmit commands to multiple underwater vehicles.
    This capability directly supports the advancement of cooperative and swarm-based operations, which are increasingly essential for large-scale and complex underwater missions.
  \item \textbf{Promoting open collaboration in marine technology.}
    In a field often constrained by high development costs and limited experimental accessibility, the open-source software release and reproducible hardware architecture foster a more dynamic and innovative community in marine robotics and acoustics.
    Researchers can extend, adapt, and customize Raspi$^2$USBL to meet specific requirements, while contributing improvements back to the broader community.
\end{itemize}

In summary, Raspi$^2$USBL not only provides a practical tool for experimental research but also lays the foundation for broader innovation and collaboration in underwater robotics and acoustics.

The remainder of this article is organized as follows. Section \ref{sec_related_work} reviews the development of the piUSBL approach and compares existing open-source acoustic systems. Section \ref{sec_system_overview} presents an overview of the Raspi$^2$USBL system, including its hardware architecture and software framework. Section \ref{sec_software_algorithm_design} details the software implementation and algorithm design. Section \ref{sec_experiments_and_analysis} describes the anechoic tank, lake, and sea experiments, along with the analysis of the results. Finally, Section \ref{sec_conclusion_future_work} concludes the article and outlines potential directions for future research.

\section{Related Work}
\label{sec_related_work}

\begin{table*}[!t]
  \caption{Comparison of Related Acoustic Systems}
  \label{tab_related_work}
  \centering
  \tiny
  \setlength{\tabcolsep}{1.0pt}
  \renewcommand{\arraystretch}{1.35}
  \resizebox{0.92\linewidth}{!}{%
  \begin{tabularx}{\linewidth}{>{\raggedright\arraybackslash}X >{\raggedright\arraybackslash}X >{\centering\arraybackslash}X >{\centering\arraybackslash}X >{\centering\arraybackslash}p{1.45cm} >{\centering\arraybackslash}p{1.45cm} >{\centering\arraybackslash}p{1.55cm}}
    \toprule
    \multicolumn{1}{c}{\textbf{Name}} & \multicolumn{1}{c}{\textbf{Affiliation}} & \makecell[c]{\textbf{System}\\\textbf{Type}} & \makecell[c]{\textbf{Open}\\\textbf{Sourced}} & \makecell[c]{\textbf{Hardware}\\\textbf{Access}} & \makecell[c]{\textbf{Software}\\\textbf{Access}} & \makecell[c]{\textbf{Raw Data}\\\textbf{Access}} \\
    \midrule
    WAYU \cite{UcnlWAYUUnderwater} & \makecell[l]{Underwater Communication \\ \& Navigation Laboratory} & LBL & \makecell{Host Computer \\ Open Sourced} & $\times$ & $\times$ & $\times$ \\
    Water Linked UGPS \cite{WaterlinkedExamplesWater} & Water Linked & SBL/USBL & $\times$ & $\times$ & $\times$ & $\times$ \\
    USBL-Acoustic-ACSM \cite{gracaRelativeAcousticLocalization2020} & University of Porto & USBL & $\checkmark$ & $\times$ & $\checkmark$ & $\times$ \\
    Underwater Acoustic.jl \cite{chitreDifferentiableOceanAcoustic2023} & \makecell[l]{National University \\ of Singapore} & \makecell{Simulation \\ Library} & $\checkmark$ & $\times$ & $\times$ & $\times$ \\
    BlueBuzz \cite{mayberryBlueBuzzOpenSourceAcoustic2022} & \makecell[l]{Georgia Institute \\ of Technology} & \makecell{Acoustic \\ Communication} & $\checkmark$ & $\checkmark$ & $\times$ & $\checkmark$ \\
    piUSBL \cite{rypkemaUnderwaterOutSight2019} & \makecell[l]{Massachusetts Institute \\ of Technology} & USBL & $\times$ & $\checkmark$ & $\times$ & $\times$ \\
    \textbf{\makecell[l]{Raspi$^2$USBL \\ (This Work)}} & Zhejiang University & USBL & $\checkmark$ & $\checkmark$ & $\checkmark$ & $\checkmark$ \\
    \bottomrule
  \end{tabularx}%
  }
\end{table*}

Table \ref{tab_related_work} summarizes and compares several existing open-source  acoustic systems. As illustrated, the commercial products such as WAYU \cite{UcnlWAYUUnderwater} and Water Linked UGPS \cite{WaterlinkedExamplesWater}, are closed-source solutions that do not provide access to hardware, software, or raw data, thereby limiting their adaptability for research purposes.
Some academic projects, such as USBL-Acoustic-ACSM \cite{gracaRelativeAcousticLocalization2020}, offer access to the processing software but lack hardware interfaces and raw data availability, which restricts their usefulness for experimental validation and customization.
The Underwater Acoustic.jl \cite{chitreDifferentiableOceanAcoustic2023} project provides a simulation library for underwater acoustics but excludes hardware or software implementations, limiting its applicability to real-world scenarios.
BlueBuzz \cite{mayberryBlueBuzzOpenSourceAcoustic2022} is an open-source acoustic communication system that provides hardware access and raw data availability but lacks processing software, which may hinder its usability.
The piUSBL system \cite{rypkemaUnderwaterOutSight2019}, which most closely resembles this work, includes hardware designed for underwater vehicles but does not offer open-source software or raw data access, constraining its adaptability and extensibility for broader research applications.
Building upon these prior efforts, this work aims to develop a comprehensive open-source USBL system that overcomes these limitations by providing full access to hardware, software, and raw data.

Before introducing the proposed system, a brief overview of the piUSBL system is provided.
The piUSBL system was first introduced by the Woods Hole Oceanographic Institution (WHOI) and the University of Washington (UW) in 2015 \cite{jakubaFeasibilityLowpowerOneway2015}.
In 2017, researchers from the Massachusetts Institute of Technology (MIT) and WHOI further developed the system and conducted both lake and marine experiments using SandShark AUVs to validate its performance \cite{rypkemaOnewayTraveltimeInverted2017,rypkemaPassiveInvertedUltraShort2019}.
In the following years, piUSBL, mounted on several AUVs, was employed for multi-vehicle cooperative positioning and navigation tasks \cite{rypkemaUnderwaterOutSight2019,rypkemaSynchronousClockRangeAngleRelative2022}.
Subsequently, researchers from Zhejiang University explored the application of the piUSBL system in long-range positioning and docking tasks for AUVs \cite{wangDesignExperimentalResults2022, wangPassiveInvertedUltraShort2022,wangAdaptiveNoiseCancelling2023}.
Currently, the piUSBL system has been successfully deployed on various platforms, including SandShark-series AUVs, Haiying-series AUVs, and BlueROV-series remotely operated vehicles (ROVs) \cite{huangGNSSaidedInstallationError2025, rypkemaHybridLongInverted2025}.

Building upon the foundational work of piUSBL, this study aims to enhance its accessibility and adaptability for a wider range of researchers and applications, particularly in the domain of underwater robotics and AUV swarm positioning.

\section{System Overview}
\label{sec_system_overview}

In this section, an overview of the Raspi$^2$USBL system is presented, including its hardware architecture and workflow.

\begin{figure}[!t]
  \centering
  \includegraphics[width=\linewidth]{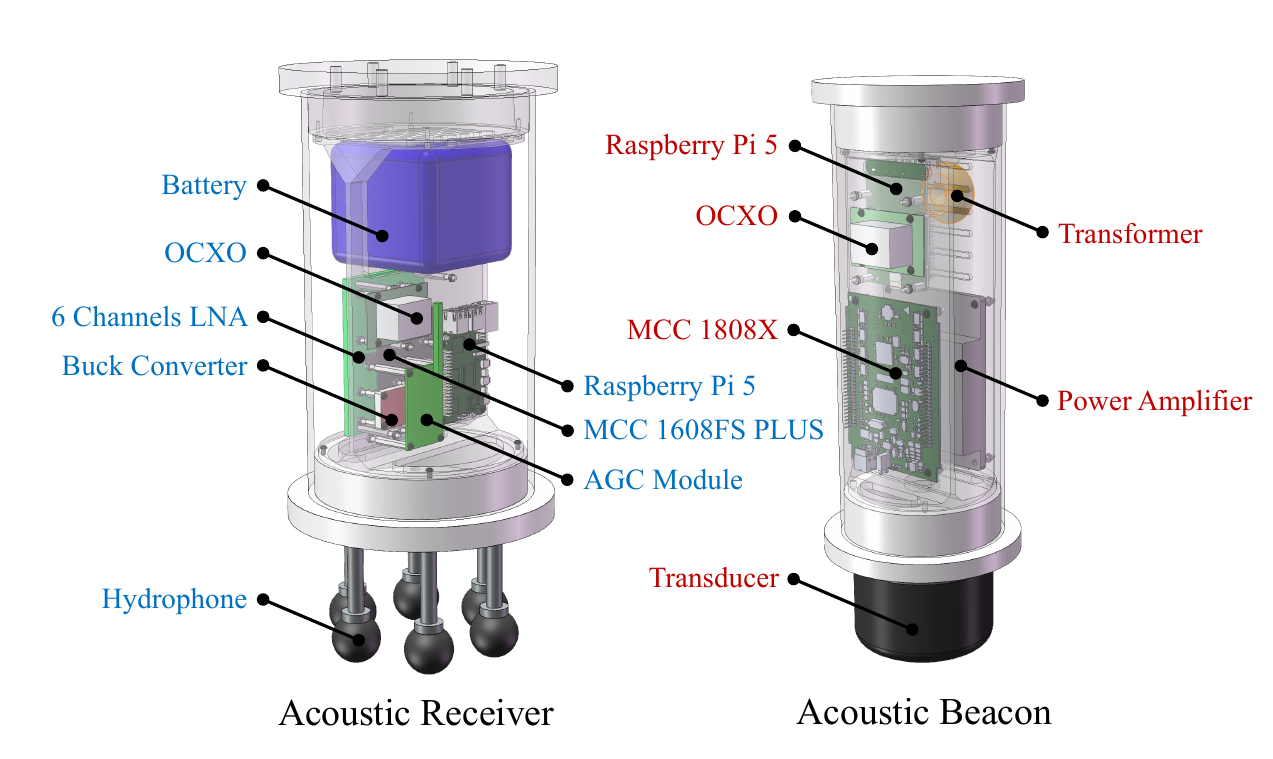}
  \caption{Hardware architecture of the acoustic receiver and beacon cabins in the Raspi$^2$USBL system.}
  \label{fig_rxtxcabin}
\end{figure}

The hardware configurations of the acoustic receiver and beacon cabin in the Raspi$^2$USBL system are shown in Fig. \ref{fig_rxtxcabin}.
The acoustic receiver cabin, designed to be mounted on underwater vehicles, provides real-time relative positioning capabilities.
It consists of a Raspberry Pi 5 as the main processing unit, a six-channel low-noise amplifier (LNA) board for acoustic signal amplification, an adaptive gain control (AGC) board for automatic gain adjustment of the LNA, a data acquisition (DAQ) board (MCC 1608FS PLUS) for analog-to-digital conversion, an oven-controlled crystal oscillator (OCXO) board for system synchronization and trigger sampling, and a 6-hydrophone array for acoustic signal reception.
The receiver cabin is powered by a 24 V battery pack, with voltage regulated to 5 V and 12 V as required by different modules.
The self-contained power architecture serves two main purposes.
First, it minimizes electrical interference from external power sources, thereby reducing noise generated by vehicle components such as thrusters.
Second, it ensures that the OCXO maintains a stable one-pulse-per-second (1 PPS) signal after disciplining, preventing clock lock loss caused by power interruptions in the AUV system, and thus maintaining continuous and reliable operation during missions.
The acoustic beacon cabin, designed to be deployed on surface platforms or leader vehicles, broadcasts acoustic signals and provides a reference position for underwater vehicles.
The beacon cabin shares a hardware architecture similar to that of the receiver cabin, except that the signal reception components are replaced with the MCC 1808X supporting digital-to-analog conversion, which serves as the digital-to-analog converter (DAC) board, a power amplifier (PA) board, a transformer for impedance matching, and a transducer for acoustic signal transmission.
Both the receiver and beacon cabins are built on the same processing platform and share a unified software framework, which streamlines system development and maintenance.

\begin{figure}[!t]
  \centering
  \includegraphics[width=\linewidth]{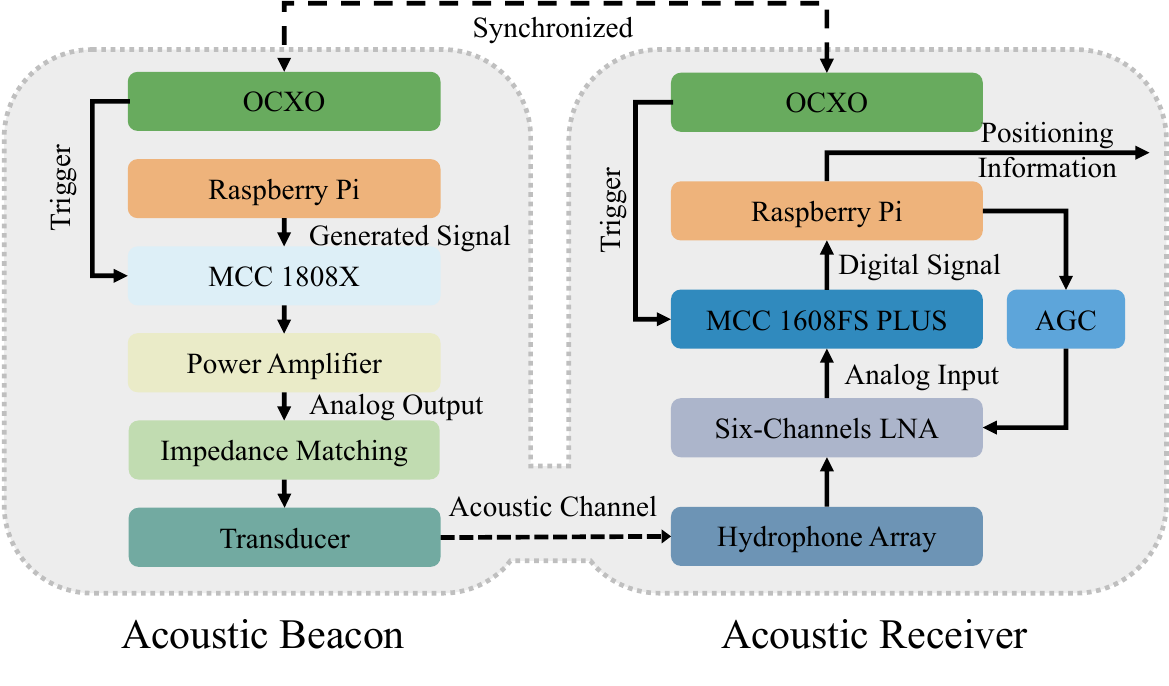}
  \caption{The system diagram of the Raspi$^2$USBL system.}
  \label{fig_systemdiagram}
\end{figure}

The system workflow is shown in Fig. \ref{fig_systemdiagram}.
Before the mission starts, the OCXOs in both the receiver and beacon cabins must be fully disciplined to the 1 PPS signal provided by a GNSS-disciplined oscillator (GNSS-DO) prior to deployment, providing an accurate and synchronized clock source for the entire system.
The disciplining process typically lasts about 30 minutes, and the OCXO (costing approximately 100 USD) can maintain a time drift of less than 180 microseconds over 12 hours.
The disciplining process is described in detail in Appendix \ref{app_time_sync_architecture}.
At a sound velocity of 1500 m/s, this drift corresponds to a ranging error of approximately 0.27 m, which is acceptable for most underwater positioning applications.

The synchronized clock is a critical component of the piUSBL system because it directly determines the accuracy and reliability of the OWTT measurements.
For the OCXOs used in this system, environmental factors such as vibration can induce frequency instability and increase clock drift over time; therefore, vibration mitigation should be considered in both the system design and deployment.
Furthermore, for long-duration and high-precision applications, two approaches can be adopted to further improve timing stability and reduce long-term drift.
First, the vehicle can periodically resurface to re-discipline the OCXO using the PPS output from the GNSS module, thereby effectively mitigating drift and maintaining synchronization during long missions.
Second, the synchronized clock can be replaced with a more stable model, or even with an atomic clock such as a chip-scale atomic clock (CSAC), to further improve timing stability at a higher cost of around USD 4000 for a drift below 100 $\mu s$/day, as discussed in related work \cite{wangDesignExperimentalResults2022}.
After the OCXOs are fully aligned, they continuously output 1 PPS signals to both the DAQ (MCC 1608FS PLUS) and DAC (MCC 1808X) boards, which trigger the sampling and transmission of acoustic signals, respectively.

With the 1 PPS signal serving as the trigger source, the Raspi$^2$USBL system supports configurable localization update rates.
The update frequency is primarily determined by the operational range and the local sound speed.
Because the ranging principle adopted in this system is based on OWTT, the theoretical minimum measurement interval is the time required for the acoustic signal to propagate from the source to the receiving array.
In underwater acoustic positioning, the update interval must be sufficiently long to accommodate the acoustic propagation time from the transmitter to the receiver.
Assuming a typical sound speed of 1500 m/s, a one-way propagation distance of 1500 m corresponds to approximately 1 s.
Therefore, for operational ranges within 1500 m, the system can operate at an update rate of 1 Hz, where each PPS signal triggers one acquisition and localization cycle.
For longer ranges, the update rate is reduced to ensure reliable signal reception.

For example, when the operational range exceeds 1500 m but remains within 3000 m, the system can be configured to perform localization every two PPS cycles, corresponding to an update rate of 0.5 Hz.
This allows sufficient time for acoustic propagation and signal processing while reducing the risk of interference between consecutive transmissions.

In practical deployments, a safety factor is introduced to account for environmental variability and system latency.
Specifically, a factor of 0.9 is applied to the maximum theoretical range.
Under a nominal sound speed of 1500 m/s, this results in a conservative threshold of approximately 1350 m for 1 Hz operation.
When the operational range exceeds this threshold, the update rate is reduced (0.5 Hz or lower) to ensure robust and reliable acoustic positioning performance.

On the beacon side, the software running on the Raspberry Pi generates the user-defined acoustic waveform and controls the DAC board to convert the digital signal into an analog waveform upon detecting the rising edge of the 1 PPS signal.
The analog signal is then amplified by the PA board and transmitted into the water through the transducer after impedance matching.

On the receiver side, the hydrophone array captures the acoustic signals and converts them into analog waveforms, which are subsequently amplified by the LNA board to ensure that the signal amplitude remains within the optimal input range of the DAQ board.
The software running on the Raspberry Pi drives the DAQ board to sample the amplified analog signals upon detection of the rising edge of the 1 PPS signal, thereby ensuring synchronization between the sampling and transmission times.
The sampled digital signals are then processed through a series of signal processing algorithms to extract the time of flight (TOF) and the direction of arrival (DOA) information, which are subsequently used to compute the relative slant range and bearing angle to the beacon.
Finally, the computed relative position is published via the serial port or Ethernet interface for other modules, such as navigation and control, to use.

Note that the hardware components, including the LNA, AGC, PA board, and transformer, are closely related to the acoustic characteristics of the hydrophones and transducers employed.
In practice, these components are typically implemented using commercial off-the-shelf modules or customized designs depending on specific deployment requirements.
To improve clarity and reproducibility, the system-level hardware architecture and functional design of key modules are further illustrated in Appendix \ref{app_hardware_architecture}.
\section{Software and Algorithm Design}
\label{sec_software_algorithm_design}

Besides the hardware design, the software implementation and algorithm design are also crucial components of the system.

\subsection{Software Design}

\begin{figure*}[!t]
  \centering
  \includegraphics[width=\linewidth]{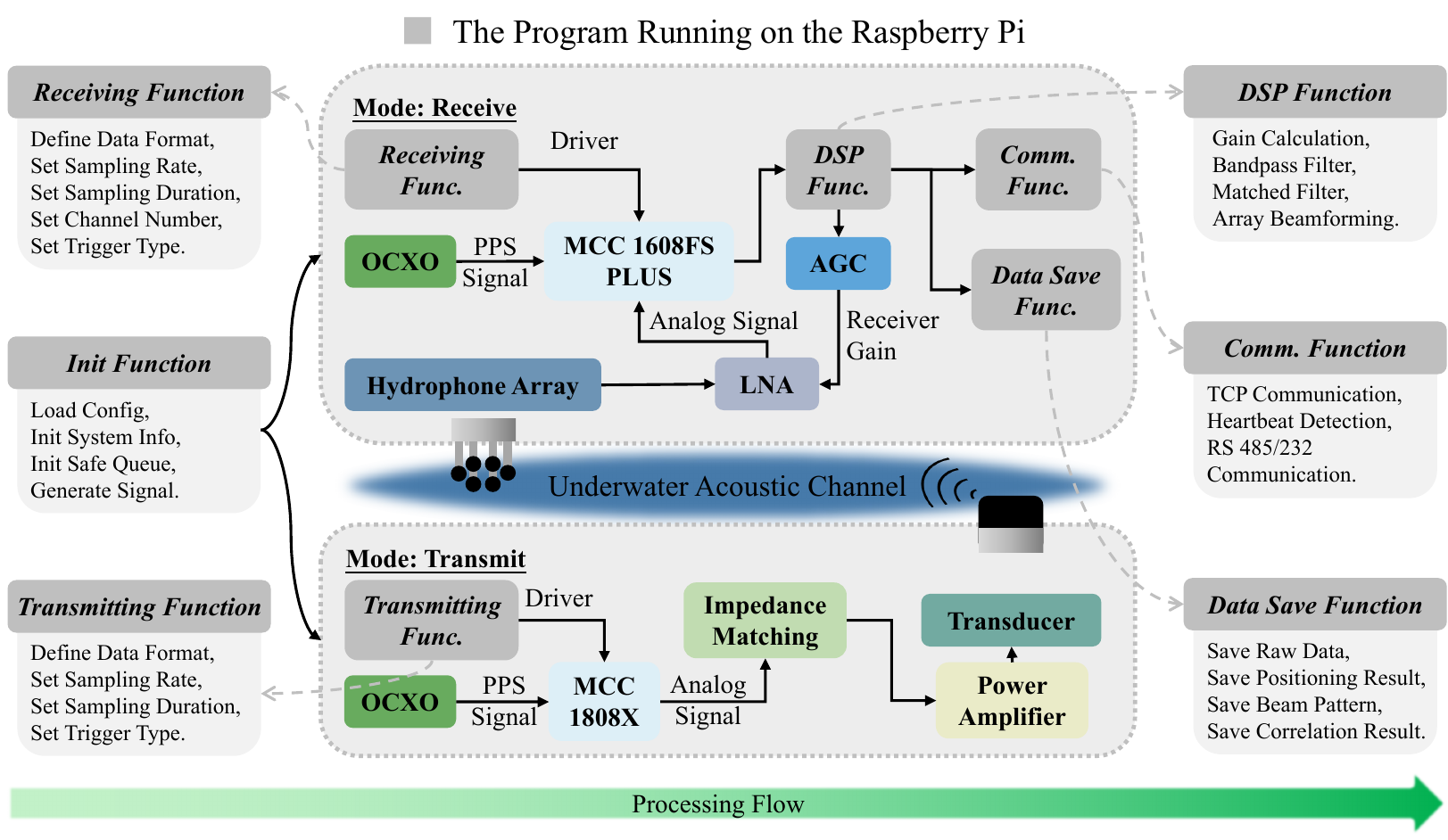}
  \caption{Software workflow of the Raspi$^2$USBL system.}
  \label{fig_softwareflow}
\end{figure*}

\begin{algorithm}[!t]
  \caption{Main processing loop of the Raspi$^2$USBL system}
  \label{alg_mainloop}
  \begin{algorithmic}
    \REQUIRE Configuration parameters from YAML file
    \ENSURE Slant range and bearing angle to the beacon
    \STATE Initialize system based on configuration parameters
    \STATE Generate reference signal $x[n]$\;
    \IF{Operating mode == Receiver}
    \STATE Initialize DAQ board
    \WHILE {System running}
    \IF {Trigger signal detected through DAQ}
    \STATE Sample multichannel signals \(\{y_i[n]\}\)
    \STATE $\hat{t} \leftarrow$ MatchedFilter$(\{y_i[n]\}, x[n])$
    \STATE $\hat{\theta} \leftarrow$ Beamforming$(\{y_i[n]\}, x[n])$
    \STATE Adaptive gain control
    \STATE Publish and log results
    \ENDIF
    \ENDWHILE
    \ELSIF{Operating mode == Beacon}
    \STATE Initialize DAC board
    \WHILE {System running}
    \IF {Trigger signal detected through DAC}
    \STATE Transmit reference signal $x[n]$
    \ENDIF
    \ENDWHILE
    \ENDIF
  \end{algorithmic}
\end{algorithm}

The software workflow is shown in Fig. \ref{fig_softwareflow} and the pseudocode of the main processing loop is presented in Algorithm \ref{alg_mainloop}.
The software is implemented in C++ and the main processing unit is a Raspberry Pi 5 running the Raspberry Pi OS, a Debian-based Linux distribution.
To ensure system integrity and facilitate maintenance, the software integrates both acoustic transmission and reception modules, with all functional components of the system logically organized within a unified framework.
Mode switching and configuration parameters, such as signal composition, transmission and sampling frequencies, duty cycle, local sound speed, and gain range, are managed through a YAML-based configuration file.
This approach enables convenient and flexible parameter adjustment according to environmental conditions during field deployments.
At initialization, the software parses the configuration file, sets the system's operating mode based on the specified parameters, generates the reference signal, and pre-allocates memory resources accordingly.

When the system operates in receiver mode, it then initializes the DAQ board (MCC 1608FS PLUS) and configures the sampling parameters, including the sampling rate, duration, channel count, and trigger source.
Then the software enters the main processing loop, continuously monitoring the received data buffer from the DAQ board, where the DAQ continuously detects the trigger signal then samples the signal from the hydrophone array.
The received data is processed in real time using a series of digital signal processing (DSP) algorithms, including bandpass filtering, matched filtering, conventional beamforming (CBF) to estimate the TOF and DOA, from which the relative position to the beacon is calculated.
The raw data, intermediate results, and final outputs are then published via the serial port or Ethernet interface for other system modules to use.
Simultaneously, the software logs all data, including raw channel data, beam pattern information, correlation results, and positioning outputs, into files for post-processing and analysis.
The overall processing pipeline is optimized for real-time execution on embedded hardware platforms.
To quantitatively evaluate the associated computational load and verify real-time performance on the Raspberry Pi platform, a detailed resource utilization analysis is presented in Appendix \ref{app_resource_utilization}.

When the system operates in beacon mode, it initializes the DAC board (MCC 1808X) and configures the transmission parameters, including the transmission frequency, signal duration, and trigger source.
The software then enters the main transmission loop, continuously monitoring the trigger signal from the OCXO and driving the DAC board to transmit acoustic signals upon detecting the rising edge.
The signal is power-amplified and transmitted into the water through the transducer after impedance matching.

\subsection{Digital Signal Processing Flow}

In this section, the key algorithms used in the Raspi$^2$USBL system are introduced, including bandpass filtering, matched filtering, conventional beamforming, and adaptive gain control.

\begin{figure*}[!t]
  \centering
  \includegraphics[width=\linewidth]{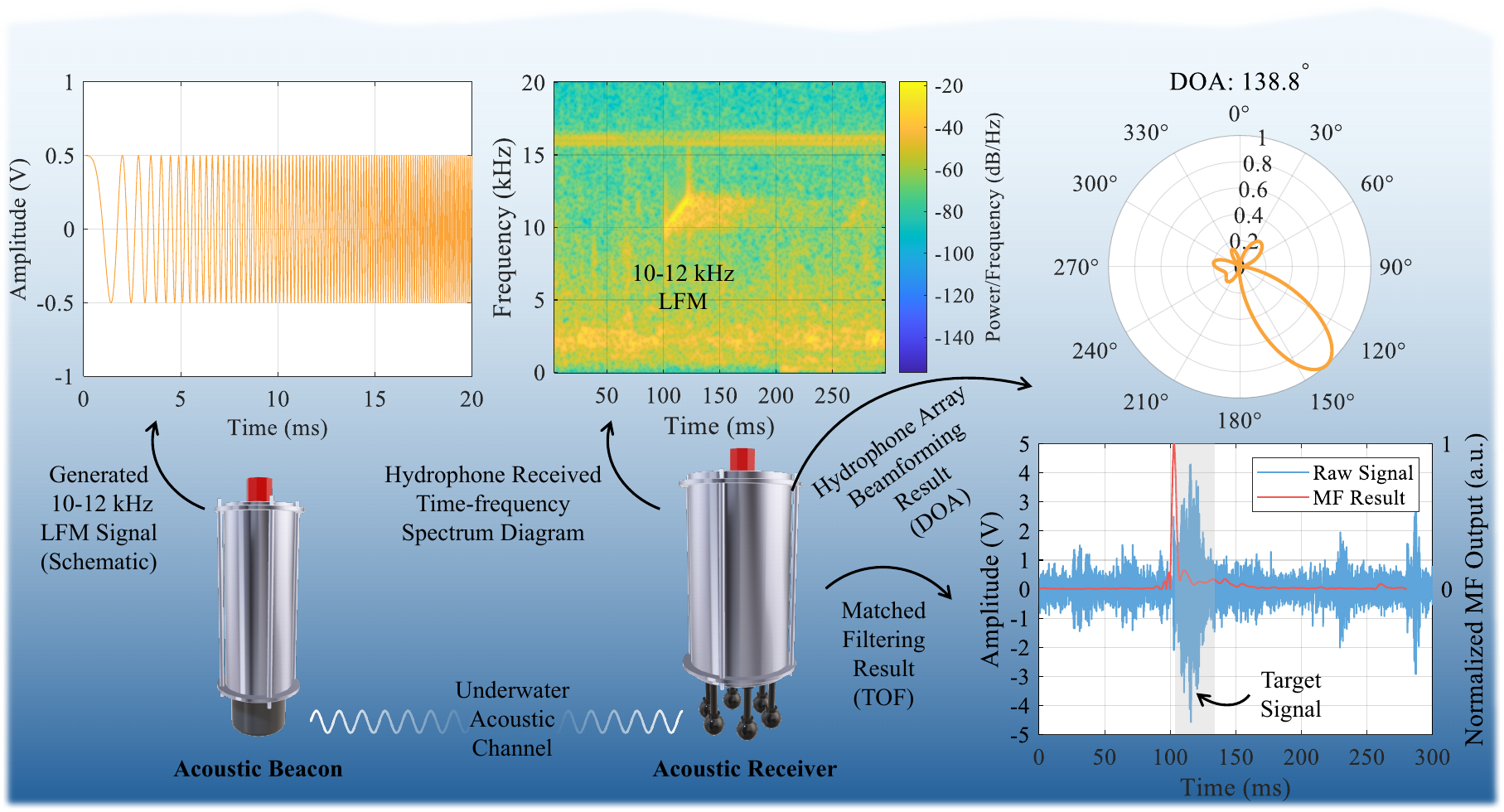}
  \caption{Signal processing workflow of the Raspi$^2$USBL system. \textit{Top Left}: The reference 10-12 kHz LFM signal used for acoustic transmission and reception. \textit{Top Middle}: The time-frequency spectrum of the received signal from one hydrophone channel. \textit{Top Right}: The beam pattern obtained from conventional beamforming for DOA estimation using the raw received signal. \textit{Bottom Right}: The matched filter output for TOF estimation and the extracted target signal segment.}
  \label{fig_signalprocessflow}
\end{figure*}

Fig. \ref{fig_signalprocessflow} illustrates the digital signal processing workflow of the Raspi$^2$USBL system, where the demonstration data are obtained from lake experiments.
In this study, a 10-12 kHz linear frequency modulated (LFM) signal is adopted as the reference waveform for acoustic transmission and reception.
Through the underwater acoustic channel, the received signal is inevitably affected by various types of noise and multipath interference, which can be clearly observed in the time--frequency spectrum of the received signal, as shown in the upper middle panel of Fig. \ref{fig_signalprocessflow}.
Therefore, the received signal first undergoes bandpass filtering to suppress out-of-band noise and interference, thereby enhancing the signal-to-noise ratio (SNR) for subsequent processing.
Following bandpass filtering, the matched filter is applied to identify correlation peaks corresponding to the transmitted signal, which helps in accurate TOF estimation for distance measurement.
The bottom-right panel of Fig. \ref{fig_signalprocessflow} shows the matched-filtering result, where the red line represents the matched-filter output, and the x-axis position of its peak corresponds to the TOF of the target signal received by this channel.
The average TOF across all channels is selected as the reference point, and a signal segment equal in length to the reference waveform is defined as the complete target signal, as the signal segment highlighted by the gray region.
The target signals from all six channels are extracted for subsequent DOA estimation, which is performed using the CBF algorithm.
The top right panel of the Fig. \ref{fig_signalprocessflow} shows the beam pattern obtained from CBF, where the polar plot depicts the signal energy distribution across different angles.
The angle corresponding to the peak of the beam pattern indicates the estimated bearing angle to the beacon.
Overall, the combination of these DSP algorithms enables the system to effectively process the received acoustic signals and accurately estimate the relative position of the beacon.

\begin{table*}[!t]
    \centering
    \caption{Unified variable definitions and their mapping to the Raspi$^2$USBL implementation}
    \label{tab_variable_definition}
    \begin{tabular}{p{0.13\linewidth}p{0.36\linewidth}p{0.39\linewidth}}
      \toprule
      \textbf{Variable} & \textbf{Definition in derivation} & \textbf{Mapping in this system} \\
      \midrule
      $A$ & Number of hydrophone elements in the array & Fixed at 6 (six-element circular hydrophone array). \\
      $R$ & Array radius & Fixed at $0.065\,\mathrm{m}$ (65 mm). \\
      $\mathbf{r}_a=[x_a,y_a]^T$ & Position vector of the $a$-th hydrophone in 2D & Computed from $(A,R)$ using uniform circular placement for $a=1,\dots,6$. \\
      $x[n]$ & Discrete reference signal & Transmitted 10--12 kHz LFM waveform generated by software. \\
      $y_i[n]$ & Received signal at channel $i$ & Digital signal sampled from the $i$-th hydrophone channel after analog front-end conditioning. \\
      $N,M$ & Lengths of $x[n]$ and $y_i[n]$ & Determined by configured signal duration and DAQ acquisition window. \\
      $f_s$ & Sampling frequency & DAQ sampling rate defined in the YAML configuration file. \\
      $t$ & TOF at array geometric center & Computed as the mean of channel-wise matched-filter peak indices divided by $f_s$. \\
      $\Delta\tau_a(\theta)$ & Relative delay of element $a$ for look angle $\theta$ & Computed from array geometry and sound speed $c$ for each steering angle. \\
      $c$ & Sound speed in water & Estimated from local environmental conditions (NPL equation) or calibrated values in controlled experiments. \\
      $f_1,f_2$ & Lower and upper frequency bounds of broadband integration & Set to 10 kHz and 12 kHz, respectively. \\
      $\theta,\hat{\theta}$ & Scanning angle and estimated bearing angle & Horizontal bearing is scanned and $\hat{\theta}$ is obtained by maximizing beam power. \\
      \bottomrule
    \end{tabular}
\end{table*}

\subsection{Matched Filter}
Table \ref{tab_variable_definition} provides a unified definition of variables used in the matched-filtering and beamforming derivations, together with their direct correspondence to the hardware and configuration parameters of the implemented six-element Raspi$^2$USBL system.

The matched filter is a classical digital signal processing technique used to estimate the correlation between a known reference signal, denoted as $x[n]$, and a received signal, $y_i[n]$, from the $i$-th hydrophone channel.
The output of the matched filter $z_i[n]$ is given by the convolution of the received signal with the time-reversed conjugate of the reference signal:

\begin{equation}
  \label{eq_zin}
  \begin{aligned}
    z_i[n] &= y_i[n] \otimes x^*[-n] \\
    & = \sum_{m=0}^{M-1} y_i[m] x^*[m-n] \\
  \end{aligned}
  \quad n \in [0, N+M-2]
\end{equation}
where $N$ denotes the length of the reference signal, $M$ denotes the length of the received signal, $\otimes$ represents the convolution operation, and $^*$ represents the complex conjugate operation.
When the received signal contains the reference signal, the output of the matched filter exhibits a distinct peak at the time index corresponding to the TOF of the reference signal.
Furthermore, owing to the circular configuration of the hydrophone array, the TOF corresponding to the geometrical center of the piUSBL system is defined as the average TOF across all channels, which can be expressed as follows:

\begin{equation}
  t = \frac{1}{6 f_s} \sum_{i=1}^{6} \arg \max_n \left( z_i[n] \right)
\end{equation}
where $t$ denotes the TOF corresponding to the geometrical center of the piUSBL system, $f_s$ represents the sampling frequency, and $\arg\max_n(\cdot)$ denotes the operator that returns the index of the maximum value.

\subsection{Conventional Beamforming}
Conventional beamforming is a widely adopted technique for estimating the DOA of acoustic signals received by a hydrophone array.
A typical CBF algorithm utilizes the array geometry and a hypothesized look angle to apply appropriate time delays to the received signals, and sums the time-delayed signals to produce an output known as the beam power \cite{wangPassiveInvertedUltraShort2022,tcyangPerformanceAnalysisSuperdirectivity2019}.
When the look angle coincides with the actual DOA of the incoming signal, the beam power achieves its maximum value.
Since AUVs are mostly equipped with depth gauge and inertial measurement sensors, only the horizontal bearing angle is estimated in this work.
This approach, when combined with TOF-based ranging, is sufficient to determine the relative position of the beacon \cite{huangPreciseTimeDelay2025}.

First, the positions of the hydrophone elements are defined in a two-dimensional (2D) Cartesian coordinate system as follows:
\begin{equation}
  \begin{aligned}
    \mathbf{r}_a & = \left[ x_a, y_a \right]^T \\
    &= \left[ R \cos \frac{2 \pi a}{A}, R \sin  \frac{2 \pi a}{A} \right]^T, \quad a = 1, 2, \ldots, 6
  \end{aligned}
\end{equation}
where $\mathbf{r}_a$ denotes the position vector of the $a$-th hydrophone element, $R$ represents the radius of the hydrophone array, and $A$ denotes the total number of hydrophone elements.
In this study, the hydrophone array consists of six elements (i.e., $A=6$), and the radius $R$ is 0.065 $\mathrm{m}$.
The relative time delay, $\Delta \tau_a$, for the $a$-th hydrophone element at a look angle $\theta$ is then calculated as follows:
\begin{equation}
  \Delta \tau_a \left(\theta\right) = \frac{1}{c} \left( x_a \cos \theta + y_a \sin \theta \right)
\end{equation}
where $c$ denotes the sound speed in water, which can be determined from the local temperature, salinity, and pressure using the National Physical Laboratory (NPL) equation \cite{leroyNewEquationAccurate2008}.

Furthermore, to enhance robustness against noise and interference, the CBF algorithm is implemented in the frequency domain.
The beam power, $B$, at a given look angle $\theta$ is computed as follows:

\begin{equation}
  B \left(\theta, f \right) =  \frac{1}{A} \sum_{a=1}^{A} X_a \left( f \right) e^{j 2 \pi f \Delta \tau_a \left(\theta\right)}
\end{equation}
where $X_a(f)$ denotes the frequency-domain representation (spectrum) of the signal received by the $a$-th hydrophone element at frequency $f$.

For a broadband signal, such as the 10-12 kHz LFM waveform used in this study, the overall beam power is obtained by integrating the frequency-domain beam power over the entire signal bandwidth, as follows:

\begin{equation}
  \label{eq:beampattern}
  B \left(\theta\right) = \int_{f_1}^{f_2} \left| B \left(\theta, f \right) \right|^2 \mathrm{d}f
\end{equation}
where $f_1$ and $f_2$ denote the lower and upper bounds of the LFM signal bandwidth, corresponding to 10 kHz and 12 kHz, respectively.

Finally, the estimated bearing angle $\hat{\theta}$, corresponding to the beacon is obtained by identifying the look angle that maximizes the beam power, as expressed by:
\begin{equation}
  \hat{\theta} = \arg \max_{\theta} B \left(\theta\right)
\end{equation}

Eq. \eqref{eq:beampattern} adopts an incoherent broadband integration strategy, in which beam power is summed across frequency bins with equal weighting. This formulation is computationally efficient and suitable for real-time embedded implementation, but it implicitly assumes that all frequency bins contribute uniformly to the final DOA estimate. In practice, underwater channels often exhibit frequency-dependent attenuation, transducer-response variation, and nonuniform ambient noise. Consequently, when signal energy and SNR are unevenly distributed across the signal bandwidth, low-SNR bins may degrade the overall DOA estimation accuracy.

Beyond this frequency-integration issue, conventional beamforming (CBF) also has an inherent resolution limitation, typically reflected in relatively broad mainlobes and elevated sidelobes. In multipath-dominant or near-field scenarios, such as shallow freshwater lakes, these characteristics may blur the direct-path peak and reduce the robustness of azimuth estimation. Increasing the array size can improve angular resolution, but this also increases hardware complexity and computational load.

High-resolution methods, such as minimum variance distortionless response (MVDR) beamforming \cite{van2002optimum} and deconvolution beamforming (dCv) \cite{tcyangPerformanceAnalysisSuperdirectivity2019,hongHighresolutionSonarImaging2025}, provide relevant alternatives by achieving narrower mainlobes and improved sidelobe suppression. Their practical performance, however, is generally more sensitive to model mismatch and array conditions. For MVDR, performance degradation may arise from structural scattering, steering-vector mismatch, sensor-gain inconsistency, and hydrophone-position perturbation. For dCv, performance is generally more favorable for approximately shift-invariant arrays with sufficiently large element counts; for the compact six-element circular array used here, violation of the shift-invariance assumption may introduce estimation bias in some look directions \cite{feng2019detection}.

Therefore, this work retains CBF with Eq. \eqref{eq:beampattern} as a baseline implementation within the proposed platform, prioritizing robustness, low implementation complexity, and stable real-time behavior on resource-constrained hardware. At the same time, coherent wideband DOA strategies and high-resolution beamforming methods, including matched-filter-domain processing and frequency-focusing approaches, will be evaluated in future extensions of the open-source software framework.

\subsection{Adaptive Gain Control}

The AGC is a signal processing technique used to automatically adjust the gain of the received signal, ensuring that its amplitude remains within an optimal range for subsequent processing.
If the signal amplitude is too low, it may be masked in noise, making detection and processing difficult.
Conversely, if the signal amplitude is too high, it may cause signal distortion and clipping, resulting in a loss of information.
Therefore, the AGC plays a crucial role in maintaining the integrity and quality of the received signal.

In this study, the maximum value of the matched-filter output is used as the signal amplitude indicator to adjust the gain of the LNA board, as defined in Eq. \ref{eq_zin}.
If this maximum value falls below a predefined lower threshold, the gain is increased by a fixed step size.
If it exceeds an upper threshold, the gain is decreased by the same fixed step size.
The thresholds and step size parameters can be configured in the system's YAML configuration file based on environmental conditions.

\section{Experiments and Analysis}
\label{sec_experiments_and_analysis}

To quantitatively assess the system's performance under both controlled and natural environments, three tiers of experiments were conducted.

\subsection{Anechoic Tank Experiment}

Through the anechoic tank experiments, the fundamental performance characteristics of the Raspi$^2$USBL system were assessed without the influence of environmental interferences such as ambient noise, multipath propagation, and Doppler shift.
Furthermore, systematic errors, including hardware delays and mechanical misalignment, were accurately characterized and compensated for by comparing the measured results against the ground truth.

\begin{figure}[!t]
  \centering
  \includegraphics[width=\linewidth]{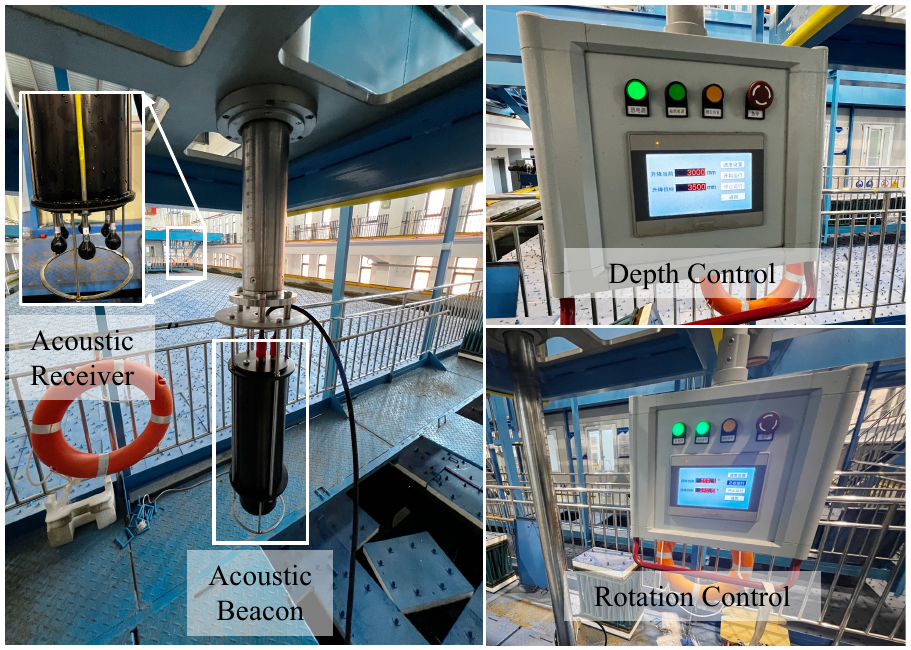}
  \caption{Anechoic tank experiment setup. \textit{Left}: The acoustic beacon cabin and the acoustic receiver cabin are both mounted on overhead traveling crane's guide columns. \textit{Right Top}: The depth control panel of the overhead traveling crane. \textit{Right Bottom}: The rotation control panel of the overhead traveling crane.}
  \label{fig_field_anechoic}
\end{figure}

\begin{table*}[!ht]
  \centering
  \caption{Specifications of the traveling crane used in the anechoic tank experiment}
  \label{tab_crane_spec}
  \begin{tabular}{lcc}
    \hline
    \textbf{Parameter} & \textbf{Range} & \textbf{Cumulative Accuracy} \\
    \hline
    Longitudinal motion (X-axis) & 50 m & $\pm$ 5 mm \\
    Transverse motion (Y-axis)   & 15 m & $\pm$ 3 mm \\
    Vertical motion (Z-axis)     & 7 m    & $\pm$ 1 mm \\
    Rotational motion            & 0 $\sim$  $\pm$ 370$^\circ$ & $\pm$0.1$^\circ$ \\
    Maximum payload              & \multicolumn{2}{c}{1 t} \\
    \hline
  \end{tabular}
\end{table*}

The experimental setup of the anechoic tank is shown in Fig. \ref{fig_field_anechoic}.
Both the acoustic beacon and the acoustic receiver were mounted on the guide columns of an overhead traveling crane, allowing precise control of their relative positions and orientations.
During the experiments, the relative distance and bearing angle between the beacon and receiver cabins were systematically varied from 1 m to 23 m in distance at 1 m intervals, and from 0$^\circ$ to 360$^\circ$ in bearing angle at 10$^\circ$ intervals.
The specifications of the traveling crane are summarized in Table \ref{tab_crane_spec}. The crane provides a longitudinal motion range of 50 m with a cumulative accuracy of $\pm$5 mm, a transverse motion range of 15 m with a cumulative accuracy of $\pm$3 mm, a vertical motion range of 7 m with a cumulative accuracy of $\pm$1 mm, and a rotational motion range of $0\sim\pm370^\circ$ with a cumulative accuracy of $\pm0.1^\circ$.
The effective sound speed in the anechoic tank was estimated from the measured TOF and known propagation distance, yielding a value of approximately 1465.8 m/s under the experimental conditions (water temperature of 15$^\circ$C and water depth of 3 m).

\begin{figure}[!t]
  \centering
  \includegraphics[width=\linewidth]{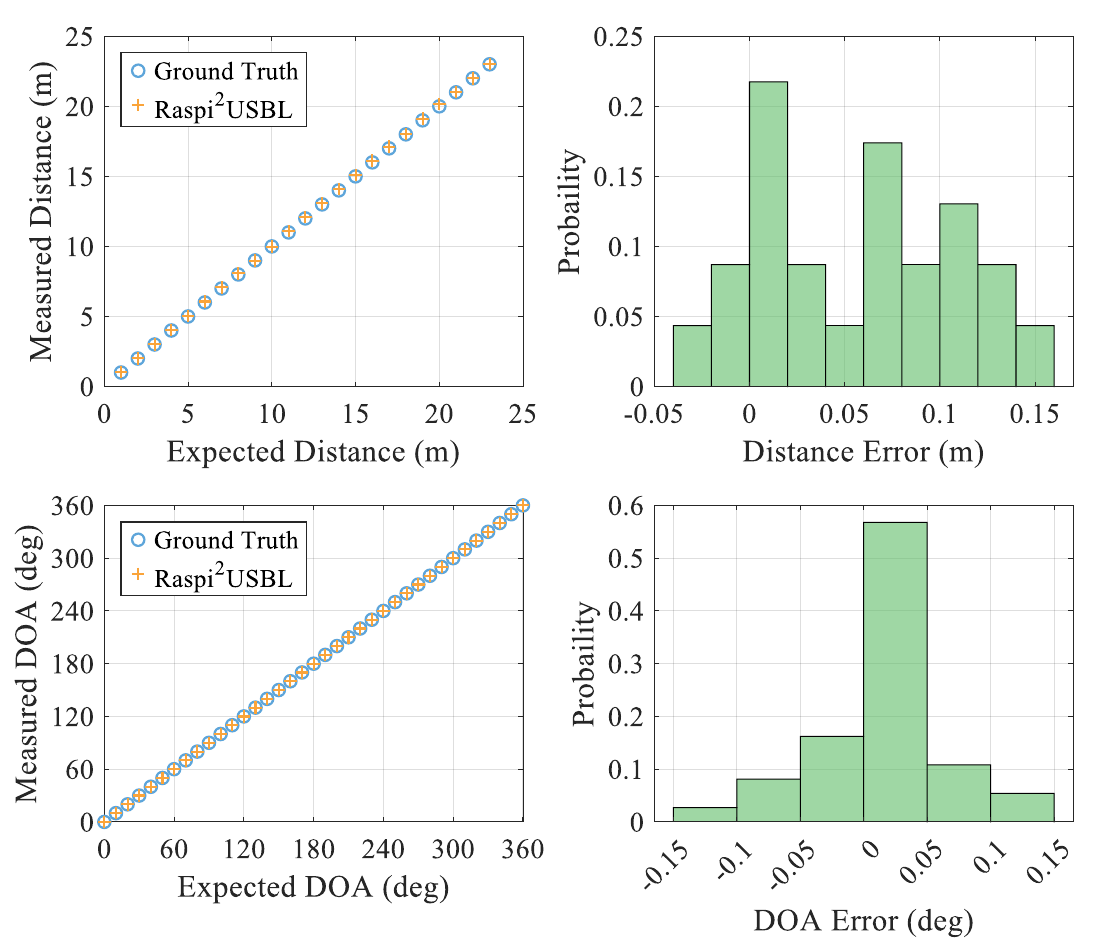}
  \caption{Experimental results of the anechoic tank experiment. \textit{Top Left}: Distance measurement results compared with ground truth. \textit{Top Right}: Error distribution of the distance measurements. \textit{Bottom Left}: DOA estimation results compared with ground truth. \textit{Bottom Right}: Error distribution of the DOA estimates.}
  \label{fig_exp_anechoic_res}
\end{figure}

The results of the anechoic tank experiments are shown in Fig. \ref{fig_exp_anechoic_res}.
The findings demonstrate that the system achieves high accuracy in both range and bearing estimation.
Compared with the ground-truth data, the root mean square error (RMSE) of the distance measurement is 0.07 m, and the RMSE of the average DOA estimation error is 0.05$^\circ$.
After applying systematic error compensation, the system achieved a ranging accuracy better than 0.1\% of the slant range and an angular accuracy of approximately 0.1$^\circ$.

\subsection{Lake Experiment}

The lake experiments were conducted to evaluate the performance of the Raspi$^2$USBL system under more realistic field conditions.
Unlike the controlled anechoic tank environment, the lake setting introduces additional challenges, including multipath propagation, ambient environmental noise, and thruster-induced interference, all of which can degrade the accuracy and reliability of the acoustic positioning system.

\begin{figure}[!t]
  \centering
  \includegraphics[width=\linewidth]{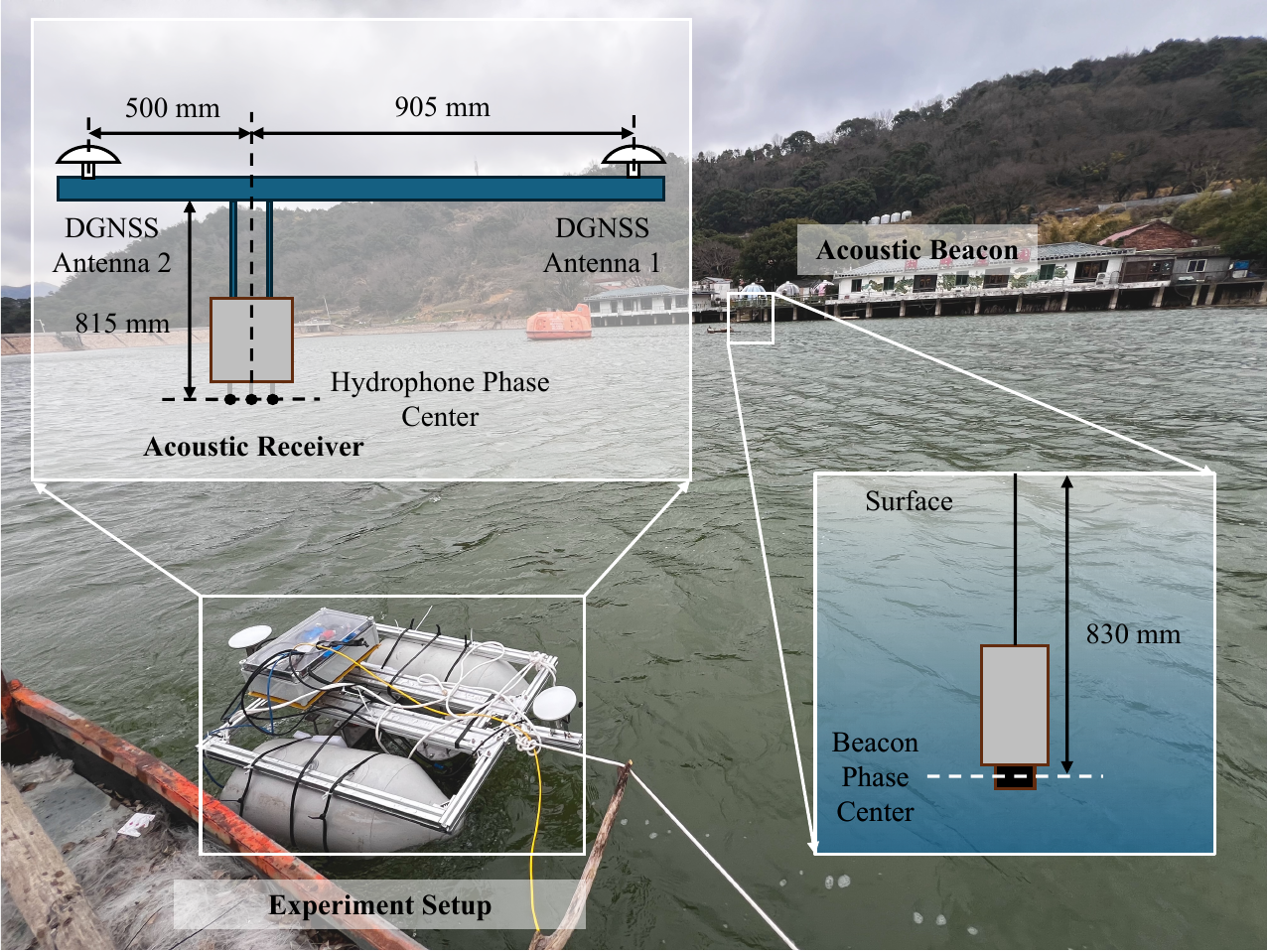}
  \caption{Lake experiment setup. \textit{Left Top}: The acoustic receiver and differential GNSS antenna mounted on the experimental setup for high-precision relative coordinate measurement. \textit{Right Bottom}: The acoustic beacon deployed underwater.}
  \label{fig_field_lake}
\end{figure}

The experimental setup of the lake tests is illustrated in Fig. \ref{fig_field_lake}, where the setup consists of an acoustic receiver cabin and a differential GNSS (DGNSS) antenna mounted on an experimental platform equipped for high-precision relative coordinate measurement, and an acoustic beacon cabin deployed underwater near the lake shore.
During the lake experiments, the water temperature was measured at 18$^\circ$C, with negligible salinity. The hydrophone array was deployed at a depth of approximately 0.5 m below the water surface. Under these conditions, the sound speed can be estimated using the NPL equation \cite{leroyNewEquationAccurate2008}.

\begin{figure}[!t]
  \centering
  \includegraphics[width=\linewidth]{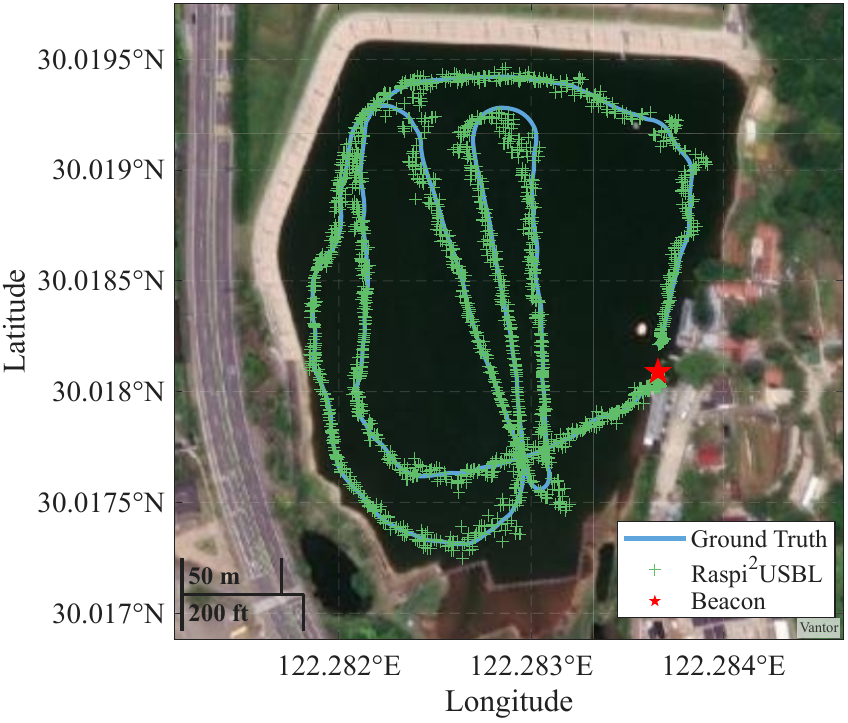}
  \caption{Results of the lake experiment. The red star marker indicates the position of the acoustic beacon, the green markers represent the acoustic measurements obtained using the acoustic receiver, and the blue line shows the DGNSS ground-truth trajectory. (Satellite imagery courtesy of Vantor).}
  \label{fig_exp_lake_res}
\end{figure}

The results of the lake experiments are presented in Fig. \ref{fig_exp_lake_res}.
In these experiments, the ground truth positions were provided by the DGNSS, while the green markers represent the acoustic measurements obtained using the Raspi$^2$USBL system.
As shown in the satellite map, the acoustically measured positions closely follow the DGNSS ground truth, demonstrating the effectiveness and reliability of the system in a realistic lake environment.
In practical applications, the vehicle depth can be obtained from an onboard depth gauge, and the relative position to the beacon can be projected onto the horizontal plane for navigation and control purposes.
During the lake experiments, the acoustic beacon was fixed at a depth of 0.83 m below the water surface.
It was assumed that the water surface remained at a uniform elevation throughout the experimental area.
The absolute elevation of the hydrophone array was determined from the measured draft of the experimental platform and the known mounting geometry of the hydrophones within the setup.

\begin{figure}[!ht]
  \centering
  \includegraphics[width=\linewidth]{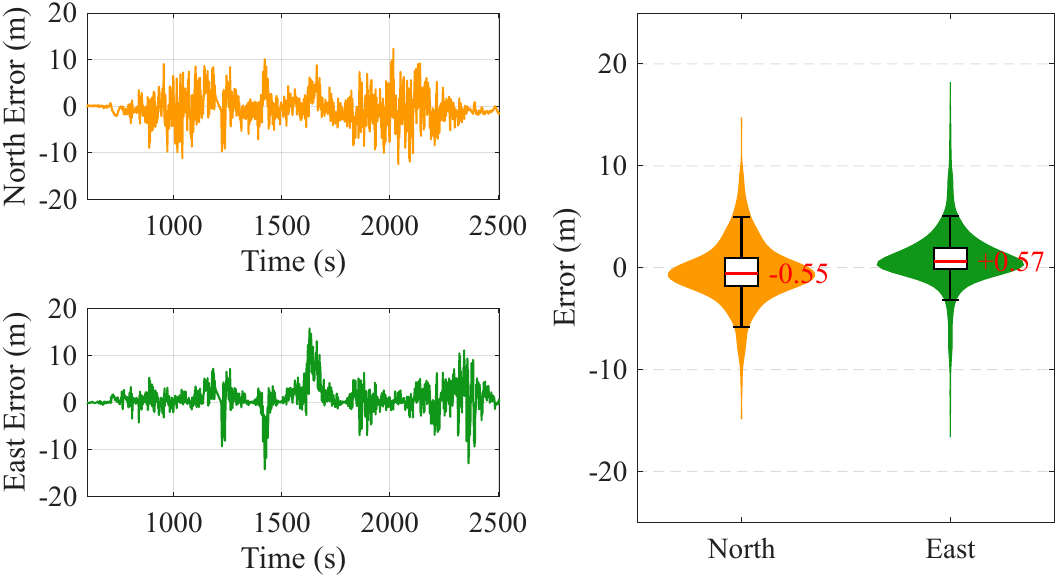}
  \caption{Positioning error of the lake experiment in the horizontal plane. \textit{Left Top}: Northward error over time. \textit{Left Bottom}: Eastward error over time. \textit{Right}: Violin plot of horizontal positioning error distribution.}
  \label{fig_exp_lake_err}
\end{figure}

Fig. \ref{fig_exp_lake_err} presents the horizontal positioning errors.
The results indicate a horizontal positioning RMSE of 4.29 m. Specifically, the RMSEs for the north and east components are 3.08 m and 2.99 m, with median errors of -0.55 m and 0.57 m, respectively.
As illustrated by the violin plot, both north and east errors exhibit approximately symmetric distributions around zero, indicating no significant systematic bias in either direction.
Most measurements fall within $\pm$5 m, demonstrating that the system performs consistently and reliably in the lake environment.
However, a few outliers with larger deviations were observed, likely caused by multipath propagation, reverberation, ambient noise, or thruster-induced interference during the experiments.
These larger deviations were particularly noticeable near the northern and southern shorelines, where shallower water depths led to stronger multipath effects.

\subsection{Sea Trial}

The sea trial was conducted in a coastal marine environment to further validate the performance of the Raspi$^2$USBL system under real-world conditions, especially for long-range underwater positioning applications in marine robotics.

\begin{figure}[!t]
  \centering
  \includegraphics[width=\linewidth]{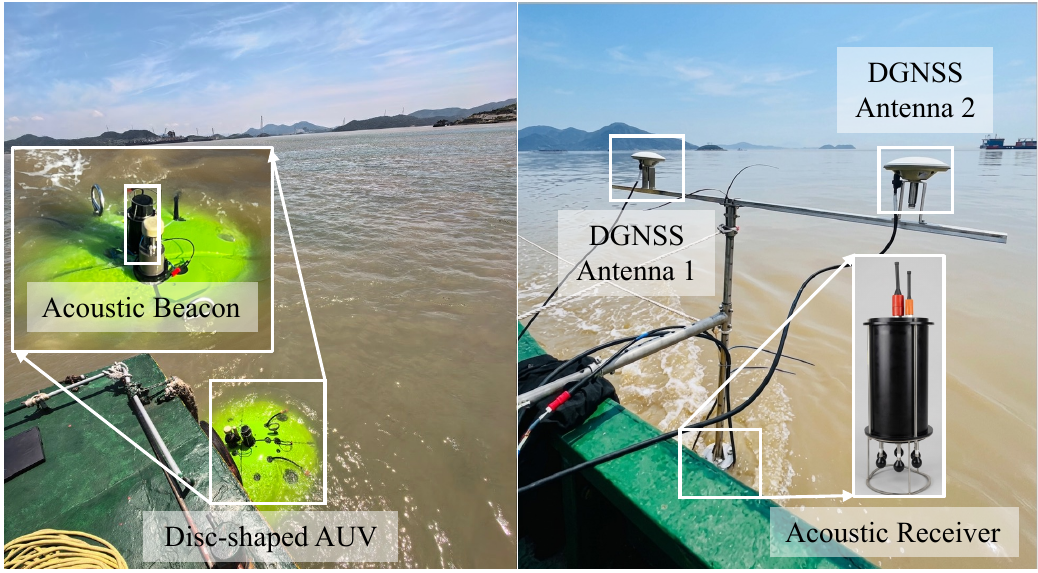}
  \caption{Setup of the sea experiment. \textit{Left}: The disc-shaped AUV equipped with the acoustic beacon cabin. \textit{Right}: The surface vessel carrying the acoustic receiver cabin and a DGNSS. }
  \label{fig_field_sea}
\end{figure}

Fig. \ref{fig_field_sea} illustrates the sea experiment setup, where the acoustic beacon is mounted on a disc-shaped AUV, and the acoustic receiver is installed on a surface vessel.
The acoustic receiver cabin is integrated with a DGNSS to provide the heading and position information for the  geometric projection of the relative position to the beacon.
It should be noted that the system installation differs from the conventional piUSBL configuration, where the acoustic receiver is typically mounted on the underwater vehicle and the acoustic beacon on the surface platform.
In this sea trial, the Raspi$^2$USBL system was configured inversely, with the acoustic receiver placed on the surface vessel and the beacon attached to the AUV, to enable real-time tracking and positioning.
This arrangement ensured reliable signal reception and processing performance on the surface and helped prevent potential loss of the AUV during the experiment.
During the experiment, the AUV descended to a predetermined depth and navigated toward the designated target point, while the acoustic positioning system onboard the vessel continuously tracked the AUV's underwater position in real time.

For the sea trials, the water temperature was approximately 10$^\circ$C, with a nominal salinity of 3.5\%, and the AUV operated at a mean depth of around 20 m.
Under these conditions, the NPL equation's effective depth was defined as the mean operating depth of the underwater vehicle over the measurement period.
A fully measured sound speed profile (SSP) was not collected in field experiments.
Given the limited depth variation and the absence of intentional operation across pronounced thermoclines or haloclines, an effective or nominal sound speed based on the measured environmental conditions was adopted as a practical approximation for the present system-level validation.

\begin{figure}[!t]
  \centering
  \includegraphics[width=\linewidth]{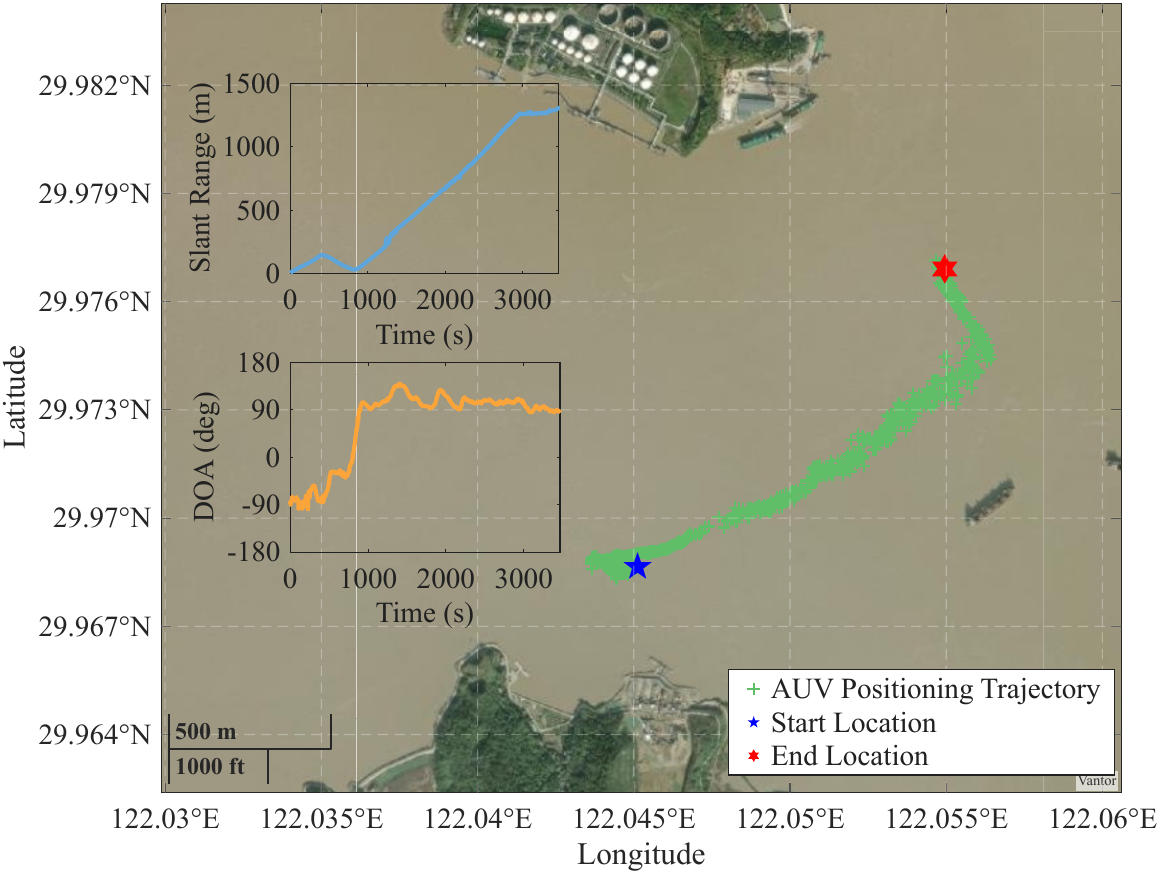}
  \caption{Results of the sea experiment. \textit{Main Panel}:  AUV trajectory tracked by the acoustic positioning system overlaid on the satellite imagery.
  \textit{Top Left Inset}: Slant-range measurements from the sea trial. \textit{Bottom Left Inset}: DOA measurements from the sea trial. (Satellite imagery courtesy of Vantor).}
  \label{fig_exp_sea_res}
\end{figure}

The results of the sea trials are presented in Fig. \ref{fig_exp_sea_res}.
The findings indicate that the maximum positioning range achieved during the trials was approximately 1309 m, and that the slant-range and DOA measurements remained stable throughout the mission.
Since the azimuth estimated by the piUSBL system is defined as a relative angle with respect to the receiver heading, the absolute bearing to the beacon was obtained by incorporating the heading information provided by the DGNSS.
During the experiment, the AUV was operated in depth-holding mode to maintain a stable vertical position.

By combining the slant range, depth, and DOA measurements with the receiver position, the relative position between the AUV and the surface vessel was geometrically projected onto the horizontal plane, thereby enabling real-time trajectory estimation.
With the DGNSS providing accurate position and heading information for the surface vessel, these relative position estimates were further transformed into the North-East-Down (NED) coordinate frame.

Based on this geometric transformation, the AUV trajectory was reconstructed by the acoustic positioning system, as illustrated on the satellite map.
Overall, the results demonstrate the practical feasibility of long-range trajectory reconstruction under realistic marine conditions.

It is worth noting that deploying a high-precision ground-truth system, such as an LBL positioning system, involves substantial cost and operational complexity, particularly in open-sea environments.
Furthermore, the primary objective of the sea trials in this study was to validate the functionality of the system on an underwater platform and assess its feasibility for long-range operation under realistic marine conditions.
Therefore, although no ground-truth reference was available, the results are sufficient to demonstrate the practical applicability of the proposed system in real-world underwater scenarios.

\section{Conclusion and Future Work}
\label{sec_conclusion_future_work}
This paper has presented Raspi$^2$USBL, a Raspberry Pi-based piUSBL acoustic positioning system developed to provide a cost-effective and reproducible research platform for underwater robotics communities.
The system integrates modular hardware, typically including the DAC/DAQ board triggered by synchronized 1 PPS signals, a six-channel low-noise amplifier, adaptive gain control, and oven-controlled crystal oscillators, with a unified C++ software framework capable of real-time digital signal processing.
Together, these components enable accurate estimation of acoustic TOF and DOA for underwater positioning tasks.

Extensive validation in three environments, a controlled anechoic tank, a freshwater lake, and open-sea trials, confirmed that the Raspi$^2$USBL achieves high slant-range accuracy better than 0.1\% of the slant range and bearing precision within 0.1$^\circ$.
The field trials demonstrated reliable long-range tracking of an AUV up to 1.3 km, verifying the system's scalability and operational robustness under realistic marine conditions.
These results indicate that the low-cost Raspi$^2$USBL framework, with open-source software and a reproducible hardware architecture, can meet the performance requirements of most underwater robotics applications.

Beyond its technical performance, Raspi$^2$USBL contributes to the underwater robotics community as a modular and extensible platform with open-source software and a reproducible hardware architecture.
By making both the source code publicly available and hardware architecture reproducible, this work supports transparent benchmarking, cross-platform comparison, and accelerated development of new algorithms for cooperative and swarm-based underwater navigation.
The framework can be readily adapted for hybrid acoustic--optical positioning, distributed AUV networks, or educational applications in ocean engineering curricula.

Future work will focus on three directions:
First, advanced signal processing techniques such as adaptive beamforming, machine-learning-based noise cancellation, and phase-coherent correlation will be integrated to mitigate multipath and reverberation effects in complex marine environments.
Second, sensor fusion with inertial measurement units (IMU), Doppler velocity logs (DVL), and depth gauge will be explored to enhance short-term stability and long-range accuracy.
Third, the open-source framework will be extended to support multi-node acoustic networking and synchronized swarm operations, enabling large-scale cooperative positioning and distributed ocean observation.

Overall, Raspi$^2$USBL establishes a practical and accessible foundation for open ocean robotics research, bridging the gap between laboratory prototyping and field, deployable underwater positioning systems, while promoting reproducibility and collaboration within the global ocean engineering community.

\appendices
\numberwithin{figure}{section}
\renewcommand{\thefigure}{\thesection\arabic{figure}}
\section{Source Code, Raw Data, and Processing Scripts}
The source code of the Raspi$^2$USBL system is available at: \url{https://github.com/ethanjinhuang/Raspi2USBL}.
In addition, a set of raw data for reference, along with relevant parameters (number of channels, sampling frequency, data layout, etc.) and Matlab processing code, are also provided in the repository for researchers to quickly analyze the raw data collected by the Raspi$^2$USBL system.

\section{Detailed Hardware Architecture}
\label{app_hardware_architecture}

This appendix supplements Section \ref{sec_system_overview} by providing compact functional diagrams of the transmit chain, receive chain, and LNA front-end components.
To avoid repeating the main text, only implementation-relevant details that are not obvious from the system overview are highlighted here.

\subsection{Transmit Chain}
\begin{figure}[!ht]
  \centering
  \includegraphics[width=\columnwidth]{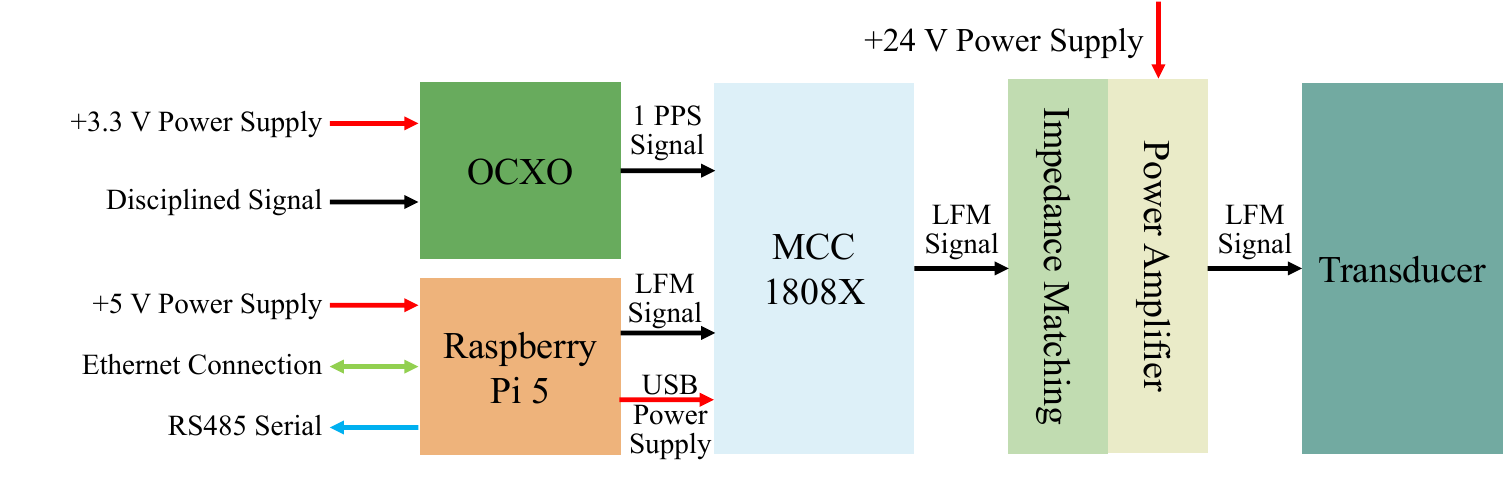}
  \caption{Transmit chain architecture of the acoustic beacon.}
  \label{fig_tx_chain}
\end{figure}

The transmit chain (Fig. \ref{fig_tx_chain}) includes a Raspberry Pi controller, a DAC module (MCC 1808X), an impedance-matching network, a power amplifier, and an acoustic transducer.

The Raspberry Pi generates the excitation signal, and the DAC outputs the analog waveform.
The PA board and matching network are selected jointly to satisfy transducer-dependent impedance characteristics in the 10--12 kHz operating band.
In practice, this matching step is critical for maintaining transmission efficiency and avoiding waveform distortion; a commercially available audio PA module is used in this implementation.
A high-stability OCXO is used as the reference clock for synchronized transmission timing.

\subsection{Receive Chain}
\begin{figure}[!ht]
  \centering
  \includegraphics[width=\columnwidth]{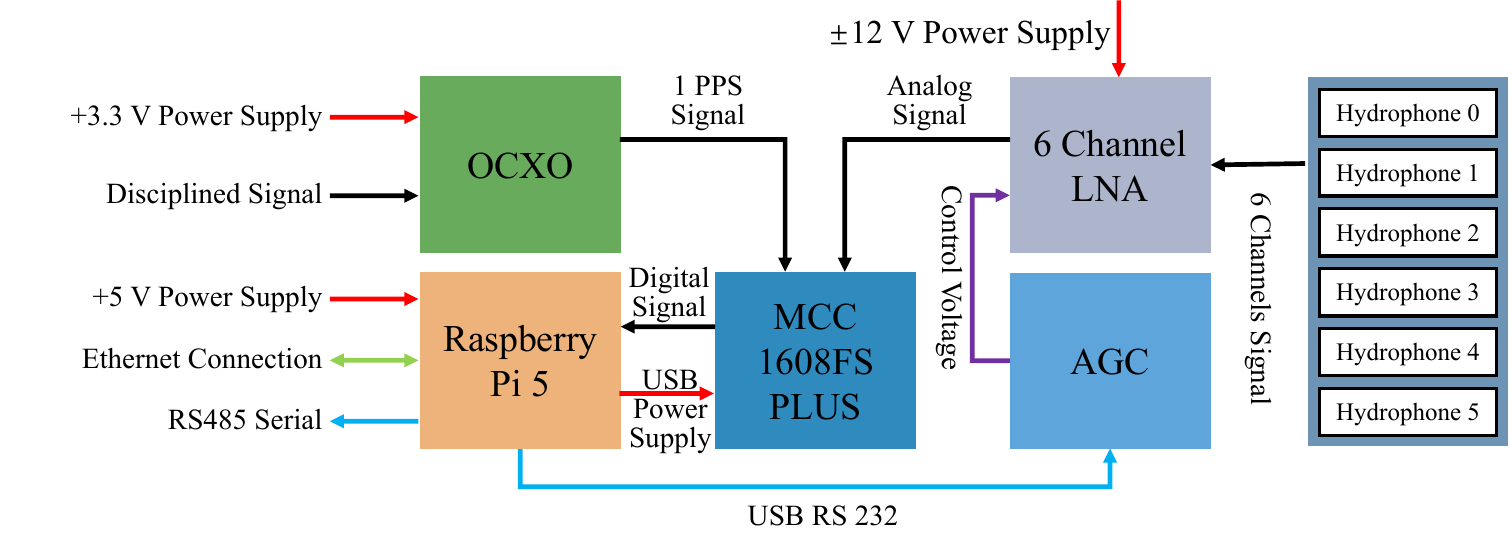}
  \caption{Receive chain architecture of the acoustic receiver.}
  \label{fig_rx_chain}
\end{figure}

The receive chain (Fig. \ref{fig_rx_chain}) consists of the hydrophone array, multi-channel LNA, AGC control path, and data acquisition module (MCC 1608FS PLUS).

The hydrophone signals are amplified by the LNA and then digitized by the DAQ for real-time processing.
AGC is implemented through a DAC-based control path: the Raspberry Pi computes the gain-control level from matched-filter outputs and updates the LNA gain in real time.
The OCXO provides the sampling-time reference required for stable multi-channel synchronization.

\subsection{LNA Module}

\begin{figure}[!ht]
  \centering
  \includegraphics[width=0.5\columnwidth]{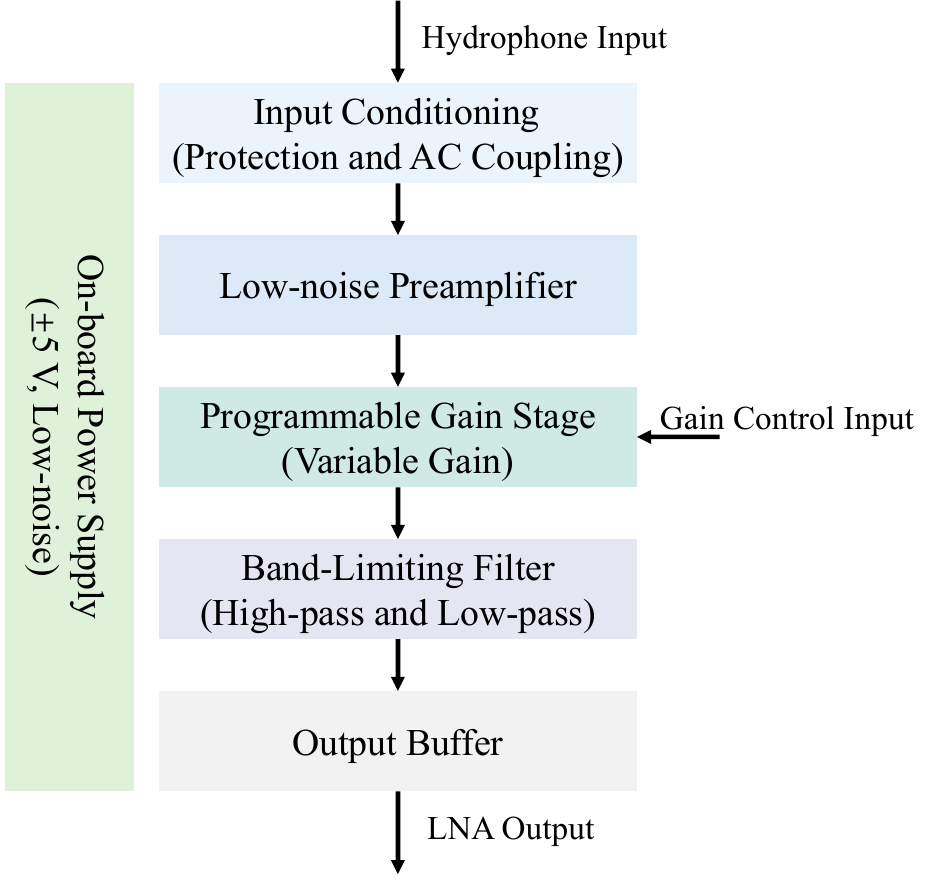}
  \caption{Functional block diagram of the LNA module.}
  \label{fig_lna_design}
\end{figure}

Fig. \ref{fig_lna_design} shows the LNA as a functional analog front-end: input protection/AC coupling, low-noise preamplification, programmable gain, band-limiting filtering, and output buffering.
The module is powered by a low-noise dual supply to maintain stable operation.

The exact LNA circuit values are hydrophone-dependent (resonance behavior, sensitivity, and admittance), so a single fixed schematic is not enforced in this paper.
Instead, this appendix provides a reproducible functional template that can be mapped to locally available hydrophones and off-the-shelf analog components.

\section{Detailed Time Synchronization Architecture}
\label{app_time_sync_architecture}
Time synchronization is a critical aspect of the Raspi$^2$USBL system, as accurate timing is essential for precise TOF estimation and coherent beamforming.
The system employs OCXOs to provide a stable and accurate time reference for both the transmitter and receiver.
The disciplining process of the OCXOs is performed prior to deployment using a GNSS-disciplined oscillator (GNSS-DO) to ensure that both the transmitter and receiver are synchronized to a common time base, even in the absence of GNSS signals during underwater operation.
For long-term and high-precision applications, the synchronized clock can be replaced with a more stable series or even the use of atomic clocks, such as a chip-scale atomic clock, to further enhance timing stability and reduce drift over extended periods, which is detailed in related works \cite{wangDesignExperimentalResults2022}.
Considering the cost and accessibility of the system, OCXOs are chosen as a practical solution for providing sufficient timing stability for most underwater positioning applications, while maintaining a low-cost and reproducible architecture.

\begin{figure}[!ht]
  \centering
  \includegraphics[width=\columnwidth]{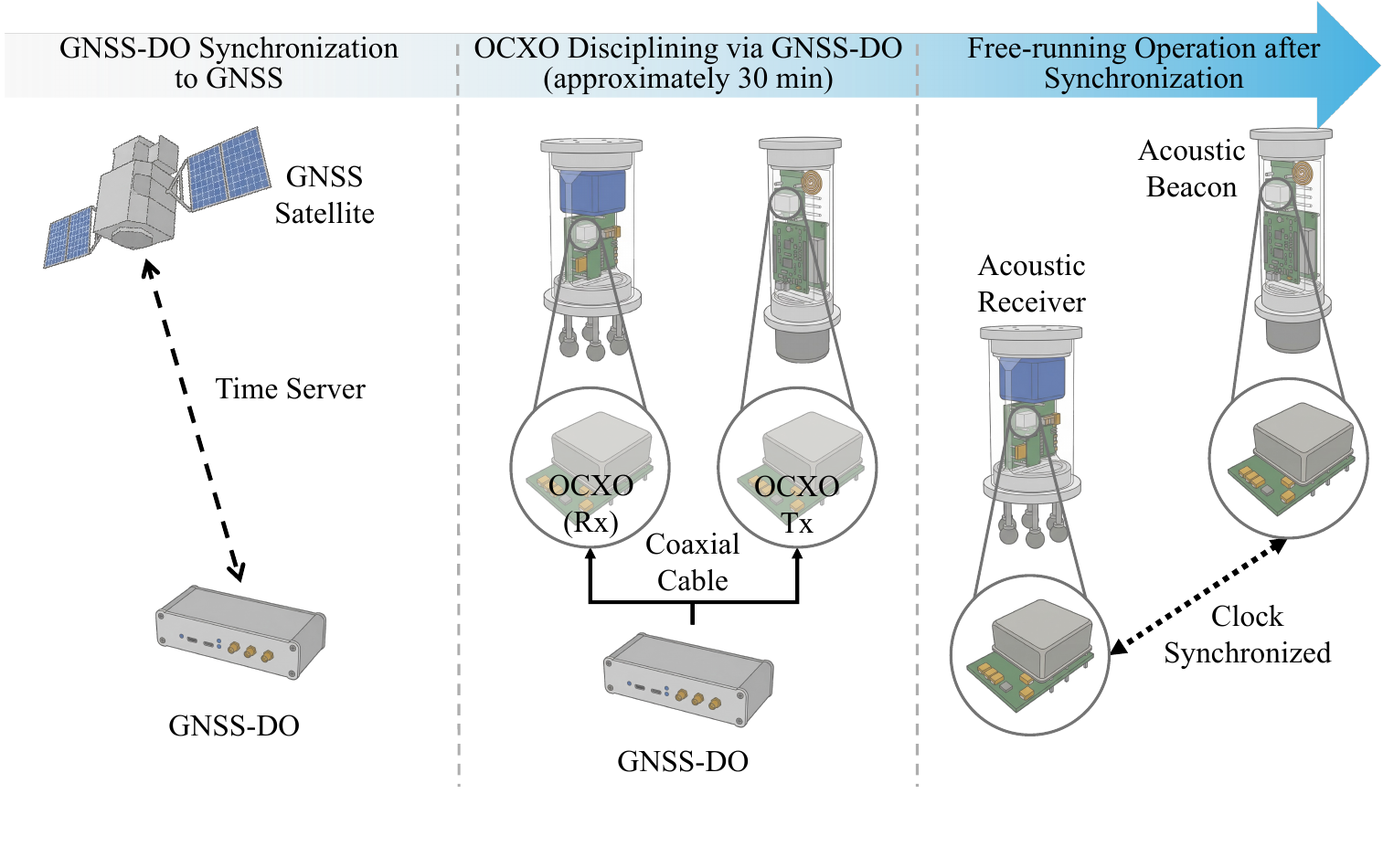}
  \caption{Clock synchronization procedure of the Raspi$^2$USBL system.
    \textit{Left}: GNSS-DO synchronized to GNSS, providing a stable time reference.
    \textit{Middle}: Pre-deployment OCXO disciplining using PPS signals distributed from the GNSS-DO via coaxial cables (approximately 30 min).
  \textit{Right}: Free-running operation after synchronization, where the transmitter (acoustic beacon) and receiver maintain timing alignment without GNSS.}
  \label{fig_time_sync}
\end{figure}

The OCXO is disciplined using the 1 PPS signal from the GNSS-DO prior to deployment.
The overall synchronization procedure is illustrated in Fig.~\ref{fig_time_sync}, which consists of three stages.

In the first stage (left), the GNSS-DO continuously tracks GNSS satellite signals and provides a high-stability time reference, including a pulse-per-second (PPS) signal. In this configuration, the GNSS-DO operates as a time server, ensuring long-term timing accuracy and stability.

In the second stage (middle), prior to system deployment, the OCXOs in both the transmitter (Tx) and receiver (Rx) are disciplined using the PPS signal distributed from the GNSS-DO via matched coaxial cables. The disciplining process is performed for approximately 30 minutes, allowing the local oscillators to converge to a common time base and minimize initial clock offset.

In the third stage (right), after the disciplining process is completed, the connection to the GNSS-DO is removed, and the system operates in a free-running mode. The transmitter (acoustic beacon) and receiver maintain synchronization based on the stability of their respective OCXOs. This approach enables reliable OWTT measurements in underwater environments where GNSS signals are unavailable.

This pre-deployment synchronization strategy eliminates the need for continuous GNSS access during operation while maintaining sufficient timing accuracy for underwater acoustic positioning.


\section{Raspberry Pi Resource Utilization}
\label{app_resource_utilization}
The run-state resource utilization of the Raspberry Pi 5 during real-time localization is summarized in Table \ref{tab_raspi_usage}.
The statistics correspond to normal runtime operation under the same configuration used in the field experiment, with a localization update rate of 1 Hz and continuous real-time processing for TOF and DOA estimation.

\begin{table}[!ht]
  \centering
  \caption{Raspberry Pi resource utilization during real-time acoustic localization}
  \label{tab_raspi_usage}
  \begin{tabular}{lll}
    \toprule
    \textbf{Metric} & \textbf{Statistic} & \textbf{Value} \\
    \midrule
    Main processing thread CPU utilization & Mean & 92.19\% \\
    Main processing thread CPU utilization & Peak & 93.40\% \\
    Localization process RSS memory usage & Mean & 29.16 MB \\
    Localization process RSS memory usage & Peak & 30.77 MB \\
    Overall system CPU utilization & Runtime & 23.68\% \\
    \bottomrule
  \end{tabular}
\end{table}

During runtime operation, the CPU utilization of the localization process remained consistently high, with a mean of 92.19\% and a peak of 93.40\%, indicating that the system made full use of the available computational resources for real-time signal processing. Despite this high utilization, the memory footprint remained modest, with a mean resident set size (RSS) of 29.16 MB and a peak of 30.77 MB, demonstrating efficient memory usage.

The overall system CPU utilization remained around 23.68\% during runtime, suggesting that the computational load was well contained within the main processing thread. The stable CPU and memory profiles confirm that the system is capable of sustained real-time operation at the configured localization update rate.

\section*{Acknowledgment}
The authors would like to express their sincere gratitude to the members of Prof. Chen's group at Zhejiang University for their valuable assistance in conducting the experiments, including Dr. Zhikun Wang, Mr. Haoda Li, Mr. Zhihang Jin, Mr. Zichen Liu, Mr. Xianyu Peng, Mr. Yuan Zhao, Mr. Yongqun Yu, and Mr. Xun Liu.
Appreciation is also extended to the staff of Zhejiang Aucean-Tech Technology Co., Ltd. for their support and collaboration during the sea trials.

\bibliographystyle{IEEEtranDOI}
\bibliography{IEEEabrv,ref}
\newpage
\begin{IEEEbiography}[{\includegraphics[width=1in,height=1.25in,clip,keepaspectratio]{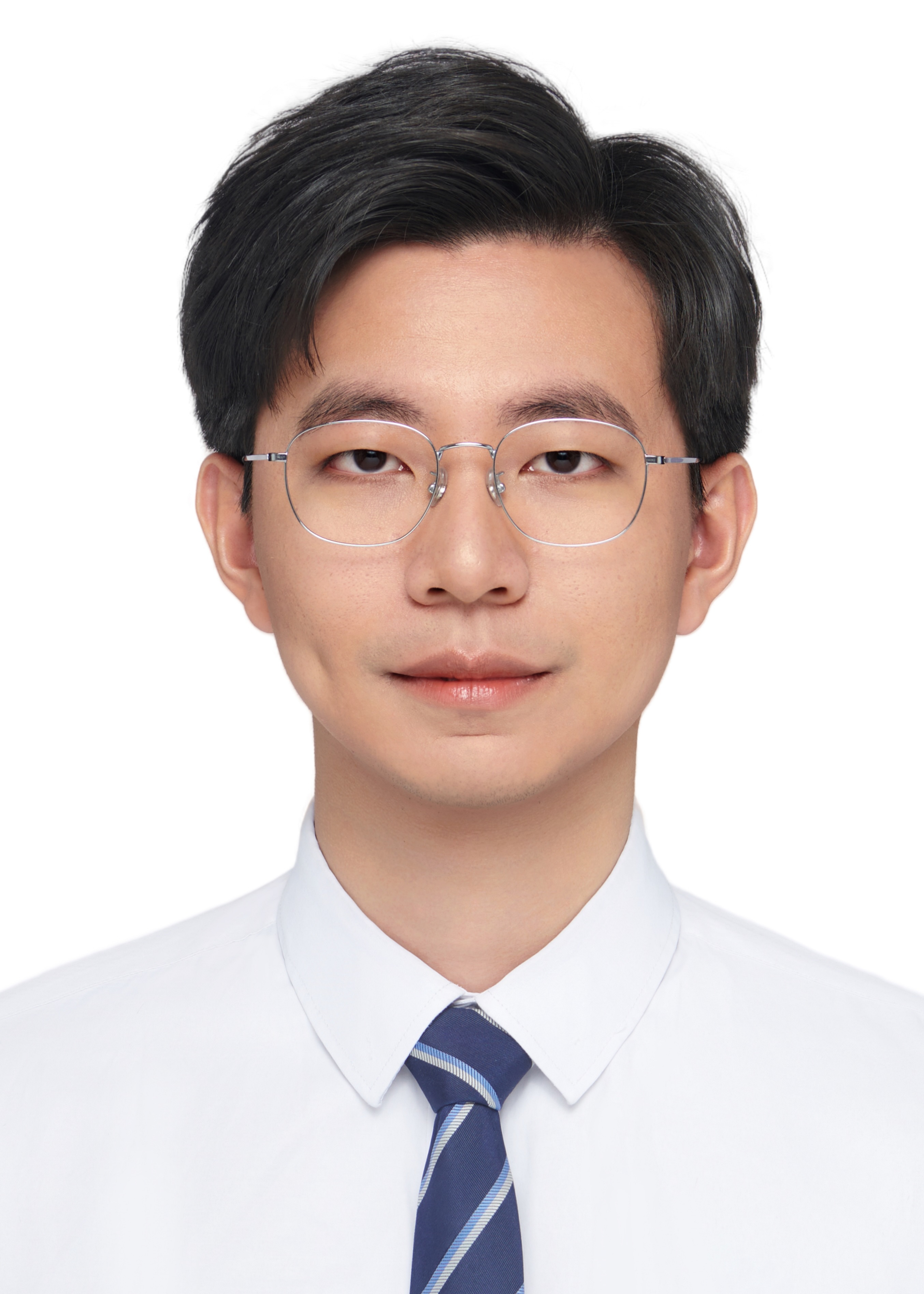}}]{Jin Huang}
  (Graduate Student Member, IEEE) was born in Qingdao, Shandong, China, in 2000. He received his B.Eng. degree in surveying and mapping engineering from Shandong University of Science and Technology, Shandong, China, in 2022. He is currently pursuing a Ph.D. degree in ocean technology and engineering at Zhejiang University, Zhejiang, China.
  \par He has been involved in the development of the piUSBL positioning system and integrated navigation systems for AUV navigation. His research interests include underwater acoustic sensing, integrated navigation, and marine robotics.
\end{IEEEbiography}

\begin{IEEEbiography}[{\includegraphics[width=1in,height=1.25in,clip,keepaspectratio]{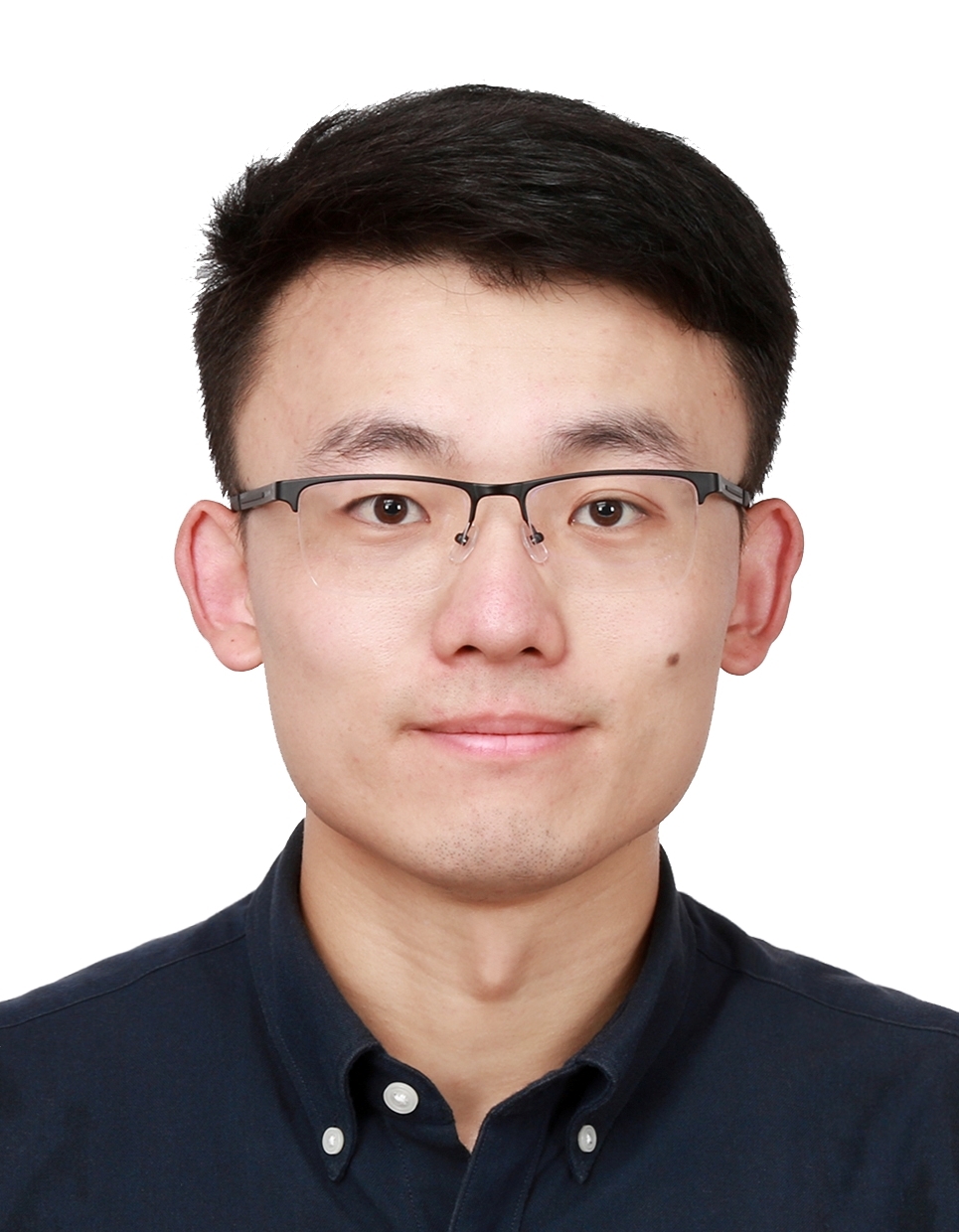}}]{Yingqiang Wang}
  (Member, IEEE) received the B.Eng. and Ph.D. degrees in ocean technology and engineering from Zhejiang University, Zhejiang, China, in 2019 and 2024, respectively. From 2022 to 2023, he was a visiting Ph.D. student with the Institute of Robotics and Intelligent Systems, ETH Z\"urich, Z\"urich, Switzerland.
  \par He is currently a Postdoctoral Research Fellow with the School of Oceanography, Shanghai Jiao Tong University, Shanghai, China. His research focuses on developing multimodal acoustic sensing, positioning, and navigation techniques to empower diverse underwater intelligent systems, such as marine robotics, with applications in deep-sea exploration.
\end{IEEEbiography}

\begin{IEEEbiography}[{\includegraphics[width=1in,height=1.25in,clip,keepaspectratio]{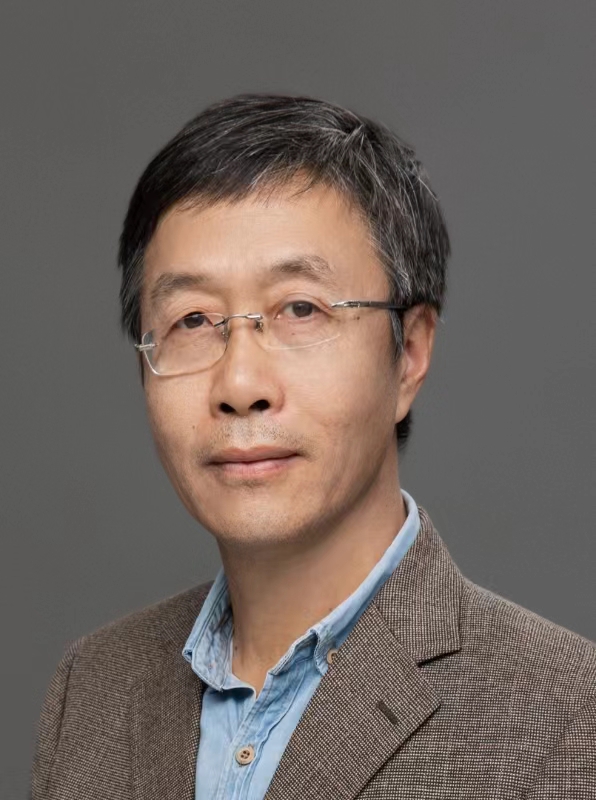}}]{Ying Chen}
  was born in Huzhou, Zhejiang, China, in 1962. He received the Ph.D. degree in mechanical engineering from Zhejiang University, Zhejiang, China, in 1989.
  \par He is currently the Founding Dean and the Qiushi Chair Professor of Ocean College, Zhejiang University. He heads a research group on deep-sea technology, which consists of technical researchers from various disciplines. His research interests include the mechatronic integration and application of deep-sea devices, many of which are related to the exploitation and observation of the deep seafloor.
\end{IEEEbiography}

\vfill

\end{document}